\documentclass[11pt]{article}

\usepackage{fontspec}
\usepackage[preprint]{acl}
\usepackage{times}
\usepackage{latexsym}
\usepackage{algorithm}
\usepackage{algpseudocode}
\usepackage{amsmath}

\usepackage[T1]{fontenc}
\usepackage[utf8]{inputenc}

\usepackage{microtype}
\usepackage{inconsolata}

\usepackage{graphicx}
\usepackage{stfloats}
 \usepackage{booktabs}
\usepackage{multirow}
\usepackage{tabularx}
\usepackage{makecell} % <-- Add this package to your preamble
\newcolumntype{C}{>{\centering\arraybackslash}X} % Define a centered column type for tabularx 

\usepackage{amssymb}
\usepackage{placeins} % in preamble

\usepackage{polyglossia}
\setmainlanguage{english}
\setotherlanguage{persian}
\newfontfamily\persianfont[Script=Arabic]{FreeSerif.otf}
\usepackage{array}
\newcolumntype{R}{>{\raggedleft\arraybackslash}X} % right-aligned column that wraps
\title{Constrained Semantic Decompression in LLMs through Persian Proverb-Conditioned Story Generation}

\author{
 \textbf{Zahra Habibzadeh\textsuperscript{2,}}\thanks{Equal contribution.}\textsuperscript{,},
 \textbf{Paria Khoshtab\textsuperscript{2,}}\footnotemark[1]\textsuperscript{,},
 \textbf{Amir Mesbah\textsuperscript{2}}\footnotemark[1]
\\, and
 \textbf{Yadollah Yaghoobzadeh\textsuperscript{1,2}}
\\
 \textsuperscript{1}Tehran Institute for Advanced Studies, Khatam University, Iran
 \\
 \textsuperscript{2}School of Electrical and Computer Engineering,
College of Engineering, University of Tehran, Tehran, Iran
\\
 \small{
   \textbf{Correspondence:} \href{mailto:z.habibzadeh213@ut.ac.ir}{z.habibzadeh213@ut.ac.ir }\href{mailto:y.yaghoobzadeh@ut.ac.ir}{y.yaghoobzadeh@ut.ac.ir}
 }
}

\begin{document}
\maketitle
\begin{abstract}
Transforming a dense, abstract proverb into an engaging and morally faithful narrative requires deep cultural understanding and robust semantic grounding. We frame this problem as a \emph{constrained semantic decompression} task and study proverb-conditioned story generation as a testbed for abstraction-to-realization in large language models (LLMs). Focusing on Persian, we introduce the Proverb Aligned Narrative Dataset (PAND), pairing proverbs with human-written stories and explicit meanings. By a hybrid evaluation framework that combines human-calibrated LLM-as-a-Judge with structural metrics, we analyze model behavior across multiple prompting regimes. Our findings reveal a persistent \emph{decompression gap}: current LLMs often achieve strong surface-level fluency while failing to faithfully instantiate the underlying moral and causal structure encoded in proverbs. We further show that explicit reasoning and iterative refinement can partially mitigate these failures, suggesting that many decompression errors arise from difficulties in translating abstract meaning into narrative form rather than a complete lack of relevant knowledge. Our proposed task naturally extends to other forms of compressed cultural knowledge.
\end{abstract}

\section{Introduction}
Stories are an important source of education and cultural transmission. Parents, teachers, and storytellers rely on narratives to communicate life lessons, social norms, and moral values. In many cultures, particularly within the Persian tradition, such narratives are closely tied to proverbs: short, memorable expressions that compress cultural wisdom, moral lessons, and traditional worldviews into condensed linguistic forms~\cite{mieder2004proverbs}.

Developing automated systems capable of understanding and generating proverb-driven narratives has both practical and scientific importance. From an application perspective, large language models (LLMs) capable of generating culturally grounded stories could support educational and heritage-preservation tools, particularly in mid-resource and low-resource languages that remain underrepresented in predominantly Western-centric training corpora~\cite{westernLLM}. More fundamentally, proverb-conditioned story generation provides a challenging testbed for evaluating whether LLMs can faithfully transform abstract cultural knowledge into coherent narrative structure.

We formalize this challenge through a \emph{constrained semantic decompression} task. Proverbs encode moral and cultural abstractions in highly compressed form; generating an appropriate story therefore requires models to transform abstract principles into concrete narrative realizations involving characters, events, and causal relations. Unlike open-ended creative writing tasks~\cite{gomez-rodriguez-williams-2023-confederacy, teleki-etal-2025-survey}, proverb-conditioned generation constrains narratives to remain faithful to the underlying moral logic rather than merely produce fluent or imaginative text. Simultaneously, the task imposes additional constraints related to narrative structure and audience appropriateness, particularly in the context of children's stories~\cite{valentini-etal-2023-automatic}.

\begin{figure*}[t!]
    \centering
    \includegraphics[width=\textwidth]{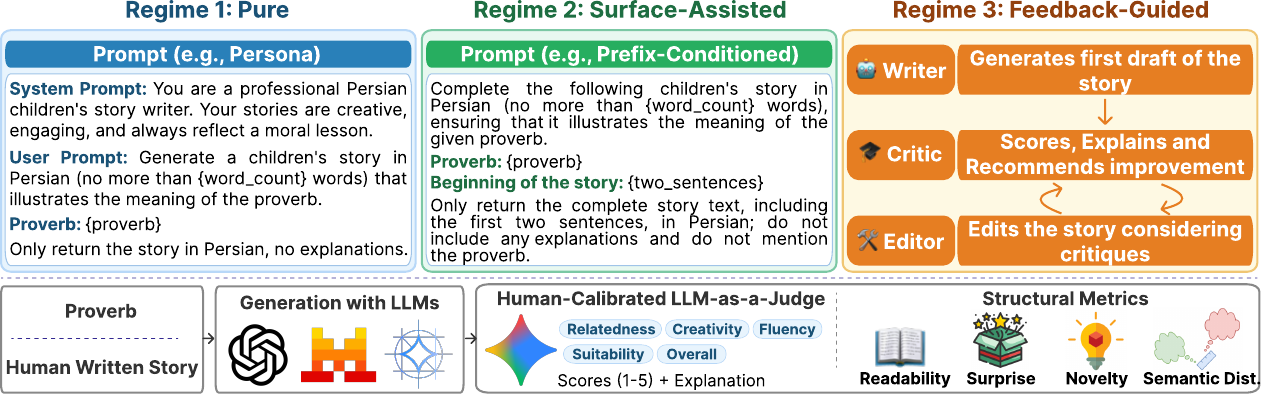}
    \vspace{-1.6em}
    \caption{The top section illustrates our decompression regimes (pure, surface-assisted, and feedback-guided) and examples of pure and surface-assisted prompts. The bottom section depicts our evaluation pipeline in which generated stories are compared to human stories via a calibrated LLM judge and structural metrics.}
    \label{fig:diagram}
\end{figure*}

To study this problem, we introduce the Proverb Aligned Narrative Dataset (PAND), a curated corpus pairing Persian proverbs with explicit meanings and human-written short stories for children. Using this resource, we evaluate a number of proprietary and smaller open-weight models across three prompting regimes illustrated in Figure~\ref{fig:diagram}: (i) \emph{Pure}, which evaluates generation and reasoning without external guidance; (ii) \emph{Surface-Assisted}, which introduces lexical cues to test reliance on surface-level signals; and (iii) \emph{Feedback-Guided}, which utilizes iterative refinement through model critique. While our experiments focus on Persian proverbs, we view semantic decompression as a broader abstraction-to-realization challenge that extends to other compressed knowledge forms, including fables, legal maxims, and moral heuristics.

We evaluate generation quality using a hybrid framework combining human-calibrated LLM-as-a-Judge with structural metrics. Specifically, judge assessments measure dimensions such as moral relatedness, creativity, and child suitability, while structural metrics quantify narrative diversity, novelty, surprise, and readability. Together, these evaluations allow us to study not only whether LLMs can generate stories from proverbs, but also how successfully they transform compressed moral knowledge into grounded narrative form faithfully.

Our results reveal a consistent decoupling between surface fluency and semantic grounding, exposing a persistent \emph{decompression gap}. While proprietary and open-weight models often generate fluent narratives, they frequently fail to preserve the deeper wisdom encoded in proverbs. We find that explicit moral reasoning improves proverb alignment, suggesting that semantic decompression benefits from separating abstraction from narrative realization. Iterative refinement further mitigates these failures, enabling smaller open-weight models to improve semantic grounding and creative narrative quality, a finding supported by a blinded human preference study.

Our contributions are as follows.
(i) We formalize proverb-conditioned story generation as a constrained semantic decompression task, providing a new framework for evaluating abstraction-to-realization capabilities in LLMs.
(ii) We introduce PAND\footnote{The dataset and code will be released publicly upon acceptance.}, the first Persian dataset pairing proverbs with explicit meanings and human-written stories for evaluating culturally grounded narrative generation.
(iii) Through extensive evaluation, we identify a decompression gap in which models achieve strong surface fluency while failing to preserve deeper moral and causal structure.
(iv) We show that explicit reasoning and iterative refinement can improve semantic grounding, enabling smaller open-weight models to approach the performance of single-pass proprietary baselines.

\section{Related Work}

\paragraph{Proverbs and Moral Reasoning:}
Prior work on proverbs and figurative expressions, such as idioms and similes, has largely focused on interpretation tasks, including detection, paraphrasing, and meaning prediction \cite{zhou2021idiomaticexpressionparaphrasingstrong, chakrabarty-etal-2022-rocket, tayyar-madabushi-etal-2022-semeval, liu-etal-2024-multilingual, magdy-etal-2025-jawaher, khoshtab-etal-2025-comparative, BANOU2025100192}. While LLMs often recover figurative meanings, recent studies show that they struggle to apply these abstractions appropriately in context, revealing a gap between abstract understanding and execution \cite{attia2025understandingevaluatingpragmaticgap}.

Research in moral reasoning has explored both moral extraction and narrative generation from abstract or visual inputs \cite{guan-etal-2022-corpus, marcuzzo-etal-2025-morables, rezapour-etal-2025-tales}. These approaches typically rely on explicit moral statements, whereas proverbs encode moral content implicitly. The closest work to our approach is \textit{ePiC} \citep{ghosh-srivastava-2022-epic},  which studies proverb-conditioned narrative generation in English using fine-tuned models. We extend this line by framing proverb-to-story generation as a \emph{constrained semantic decompression} task, requiring models to expand implicit morals into target-appropriate narratives via prompting.

\paragraph{Automatic Story Generation:}
Advances in LLMs have improved automatic story generation, with extensive work dedicated to hierarchical and planning-based approaches that separate high-level content planning from surface realization \cite{fan-etal-2018-hierarchical, fan-etal-2019-strategies, 10.1609/aaai.v33i01.33017378, goldfarb-tarrant-etal-2020-content}. Modern approaches like \textit{Re3} \cite{yang-etal-2022-re3}, \textit{Agents' Room} \cite{huot2025agents}, \textit{DOME} \cite{wang-etal-2025-generating}, and \textit{BookWorld} \cite{ran-etal-2025-bookworld} decompose generation into multi-stage or persona-based multi-agent processes.

Another prominent research branch emphasizes iterative, feedback-based refinement. \textit{Self-Refine} \cite{10.5555/3666122.3668141}  shows that an LLM can generate feedback to improve its own outputs, while methods such as \textit{CritiCS}~\cite{bae-kim-2024-collective}, \textit{SWAG}~\cite{pei-etal-2024-swag}, and \textit{Dramaturge}~\cite{xie2025plugandplaydramaturgedivideandconquerapproach} integrate external critics or structured editorial guidance. These approaches and similar work on other tasks, such as translation~\cite{chen-etal-2024-iterative}, primarily focus on evaluating proprietary models in English and target open-ended, long-form narratives. We specifically focus on abstraction-to-realization under cultural and semantic constraints in Persian. Importantly, our feedback regime exclusively utilizes open-weight models, enabling local, cost-effective deployment.

\paragraph{Story Generation Evaluation:}
Early evaluation frameworks focused on surface-level narrative properties such as structural coherence, text redundancy, and basic linguistic fluency \cite{chhun-etal-2022-human}.  Recent methodologies have shifted toward creativity-oriented benchmarks, evaluating LLM-generated narratives against human writing via human annotation and automated metrics \cite{chakrabarty-etal-2023-creative,johnson2023divergent,orwig2024language,tian-etal-2024-large-language,marco-etal-2024-pron, atmakuru2024cs4measuringcreativitylarge,marco-etal-2025-small,lu2025ai}. Despite high surface fluency, these studies uncover limitations in stylistic diversity and originality. They primarily evaluate generation in unconstrained, open-ended settings. In contrast, we treat story generation as a constrained semantic grounding problem rather than an open writing task, disentangling moral faithfulness from surface text quality. We align our subjective criteria with \textit{ePiC} \cite{ghosh-srivastava-2022-epic} and adapt structural metrics from \cite{ismayilzada2024evaluating} to Persian, enabling a robust evaluation of moral grounding alongside creative narrative expansion.

\section{Dataset}
\label{sec:dataset}
Our Persian gold dataset, called PAND, contains 150 instances. Each instance consists of a proverb, its meaning, and a human-written short children's story corresponding to that proverb. This is the first Persian resource to combine these three elements into a single, structured corpus. Detailed statistics of the PAND dataset appear in Table \ref{tab:dataset-statistics}. Table \ref{tab:dataset-example} in Appendix \ref{sec:appendix-data} presents an example of our dataset.

\begin{table}[ht]
    \centering
    \small
    \begin{tabular}{@{}lll@{}}
        \toprule
        \textbf{Category} & \textbf{Metric} & \textbf{Value} \\ 
        \midrule
        \multirow{2}{*}{\textbf{Dataset Scale}} & Total Number of Stories & 150 \\
         & Unique Proverbs & 116 \\
        \midrule
        \multirow{4}{*}{\textbf{Story Length}} & Average Words per Story & 369 \\
         & Maximum Words in a Story & 1462 \\
         & Average Sentences per Story & 22.3 \\
         & Maximum Sentences in a Story & 83 \\
        \midrule
        \multirow{2}{*}{\textbf{Proverb Length}} & Average Words per Proverb & 7.6 \\
         & Maximum Words in a Proverb & 18 \\
        \bottomrule
    \end{tabular}
    \vspace{-0.5em}
    \caption{PAND dataset statistical summary.}
    \label{tab:dataset-statistics}
\end{table}

\subsection{Data Collection}
We built this dataset by crawling public Persian literary and educational websites for pairs of Persian proverbs and children's short stories (listed in Table~\ref{tab:links}, Appendix~\ref{sec:appendix-data}).
Then, we separately collect semantic meanings for each proverb from reputable online Persian resources, including \textit{Abadis}\footnote{\href{https://abadis.ir/}{https://abadis.ir/}}, an online dictionary, and \textit{Daneshchi}\footnote{\href{https://www.daneshchi.ir/}{https://www.daneshchi.ir/}}, an educational platform. These meanings provide explicit explanations of the proverb's moral or metaphorical content. They are used in our subjective evaluation to assess the relatedness between generated stories and the intended moral meanings of the proverbs (see Section~\ref{sec:subjective_evaluation}).

\subsection{Data Preprocessing and Cleaning}
The limited size of our dataset stems from our strict quality constraints for a mid-resource language. We prioritized a high-quality, human-verified benchmark over a larger, noisy dataset. Therefore, we manually verified that every proverb and human-written story was published before the LLM era to prevent synthetic contamination. To improve data quality, we preprocess and clean the raw data by removing irrelevant proverb-story pairs, discarding stories shorter than 50 words or longer than 1500 words, normalizing punctuation, and correcting half-space and paragraph structure issues common in Persian text. We utilized LLMs for normalization instead of established rule-based tools because of their power in handling context-sensitive punctuation and inconsistencies in web crawls (Figure~\ref{fig:english-edit} in Appendix~\ref{sec:appendix-prompt}). The cleaned dataset contains 150 proverb-story pairs covering 116 unique proverbs. For proverbs with multiple stories, we retain a single reference human-written story per proverb for our experiments, selected by the highest Overall score, with ties broken by the sum of remaining metrics (see Section~\ref{sec:subjective_evaluation}).

\section{Proverb Decompression Strategies}
We formalize proverb-conditioned story generation as mapping a compressed proverb $P$ to an expanded narrative $S$. We treat this process as a \emph{constrained semantic decompression} task in which $S$ must preserve the moral intent of $P$ while satisfying narrative length and child-appropriateness constraints. Due to the scarcity of domain-specific Persian instruction data for post-training, we investigate inference-time prompting strategies that vary in the degree of external guidance. Our goal is to identify mechanisms that improve semantic grounding while preserving narrative quality.

We evaluate these strategies across several models with verified Persian proficiency~\cite{abaskohi-etal-2024-benchmarking}, including GPT-4.1~\citep{openai2024gpt4technicalreport}, Gemma~3~(12B and 27B)~\citep{gemmateam2025gemma3technicalreport}, and Mistral Small~3.2~24B~\citep{mistralModel}. Detailed model settings and prompt templates are provided in Appendix~\ref{sec:appendix-experiment} and Appendix~\ref{sec:appendix-prompt}.

\subsection{Pure}
This regime provides only the proverb and a word limit, isolating the model’s intrinsic ability to interpret and ground abstract moral meaning without external cues. We evaluate four prompting strategies: (i) \textbf{Zero-Shot}, which directly instructs the model to generate a children's story from the proverb; (ii) \textbf{Persona}, which asks the model to act as a professional Persian children's author; (iii) \textbf{Moral CoT}, a chain-of-thought approach~\cite{10.5555/3600270.3602070} in which the model explicitly identifies the proverb’s moral lesson before story generation; and (iv) \textbf{Outline CoT}, a plan-and-write strategy in which the model first generates a narrative outline based on a classical three-act structure~\cite{aristotle1902poetics} before realizing the final story.

\subsection{Surface-Assisted}

 This regime augments prompts with hints derived from human-written stories in PAND to examine how surface-level cues influence semantic grounding and generation behavior. Because these cues partially expose reference content, this regime also introduces a potential risk of memorization. We consider two prompting strategies: (i) \textbf{Prefix-Conditioned}, which provides the first two sentences of the reference story as a continuation prompt; and (ii) \textbf{Cue Words}, which supplies a small set of keywords extracted from the reference story that the model must incorporate into the generated narrative. The extraction prompt is shown in Figure~\ref{fig:english-cue} in Appendix~\ref{sec:appendix-prompt}.

\subsection{Feedback-Guided}
Writing is inherently iterative~\cite{ECKSTEIN2011162}. Motivated by human editorial workflows, we introduce a \emph{Feedback-Guided} regime in which story generation is refined through critique and revision. As illustrated in Figure~\ref{fig:diagram}, this regime decomposes generation into three personas: a \emph{Writer}, a \emph{Critic}, and an \emph{Editor}.

The process begins with the \emph{Writer} generating an initial story conditioned on the proverb. The \emph{Critic} then evaluates this draft according to the subjective criteria described in Section~\ref{sec:subjective_evaluation}, producing both scalar scores and natural-language feedback identifying abstraction or grounding failures. Finally, the \emph{Editor} revises the narrative by incorporating the critique to improve semantic alignment with the proverb. This interaction forms an iterative refinement loop that can be repeated multiple times. Prompt of each persona and implementation details are provided in Appendix~\ref{sec:appendix-prompt} and Appendix~\ref{sec:appendix-refine-detail}.

\section{Evaluation Methods}
We evaluate generated stories through a hybrid framework pairing human-calibrated LLM judgments with objective structural metrics to capture both qualitative alignment and structural creativity.

\subsection{Subjective Evaluation of Story Quality}
\label{sec:subjective_evaluation}

Evaluating semantic decompression requires assessing not only surface narrative quality but also the faithful grounding of abstract moral content. We define five subjective metrics that capture different interpretive dimensions of semantic decompression. All criteria are rated on a 5-point Likert scale. \textbf{Relatedness} measures how faithfully the generated story conveys the meaning or moral of the given proverb (1: unrelated; 5: fully faithful). \textbf{Creativity} assesses the originality and narrative interest of the story beyond literal restatement (1: repetitive or uninteresting; 5: highly creative and engaging). \textbf{Fluency} evaluates grammatical correctness and readability (1: frequent errors and poor readability; 5: grammatically correct and easy to read). \textbf{Suitability for Children} rates the appropriateness of content, language, and conveyed message (1: inappropriate; 5: highly suitable). \textbf{Overall} provides a holistic assessment of narrative quality (1: very low; 5: excellent). Except suitability for children, all other metrics are adopted from \textit{ePiC} \cite{ghosh-srivastava-2022-epic} framework.

\begin{table}[ht]
    \centering
    \resizebox{\columnwidth}{!}{%
    \begin{tabular}{l c | cccc}
        \toprule
        \multirow{2}{*}{\textbf{Metric}} & \textbf{Human--Human} & \multicolumn{4}{c}{\textbf{Gemini--Human Agreement}} \\
        \cmidrule(lr){3-6} % Only drawing the line for the LLM section now
         & \textbf{Agreement} &  \textbf{Mean} & \textbf{Median} & \textbf{All} \\
        \midrule
        Relatedness           & 0.88 & 0.74 & 0.72 & 0.87 \\
        Creativity          & 0.83 & 0.55 & 0.51 & 0.79 \\
        Fluency               & 0.48 & 0.43 & 0.38 & 0.53 \\
        Suitability for children & 0.76 & 0.68 & 0.69 & 0.78 \\
        Overall               & 0.82 & 0.72 & 0.66 & 0.83 \\
        \bottomrule
    \end{tabular}%
    }
    \vspace{-0.5em}
        \caption{Inter-annotator agreement (ICC3k) between human annotators and between Gemini 2.5 Pro and human annotators.}
    \label{tab:annotator-agreement}
\end{table}

While human evaluation offers the highest fidelity, it does not scale efficiently. To scale these evaluations efficiently while preserving cultural and linguistic nuances, we deploy a calibrated LLM-as-a-Judge~\cite{li-etal-2025-generation}. We verified the judge's reliability by conducting a calibration study comparing a diverse range of proprietary and open-weight models against a human gold standard (Appendix~\ref{sec:appendix-judge} details our study and analysis). Three graduate students annotated a control set of 40 paired human and GPT-4.1 stories (80 total) using our annotation guidelines (see Appendix~\ref{sec:appendixB}). Annotation reliability was assessed using the Intraclass Correlation Coefficient (ICC(3,k)) \cite{shrout1979intraclass}. As shown in Table~\ref{tab:annotator-agreement}, annotators exhibit strong agreement on key metrics, including relatedness (0.88) and creativity (0.83), confirming the reliability of our evaluation protocol.

Table~\ref{tab:annotator-agreement} also reports agreement metrics for our selected judge, Gemini 2.5 Pro \cite{comanici2025gemini25pushingfrontier}. Under the ``All'' aggregation setting, where the judge acts as a fourth independent annotator, Gemini 2.5 Pro achieves strong agreement with humans, with correlations of 0.87 for relatedness and 0.83 for overall quality, supporting its use as a reliable and scalable proxy for human judgment in our subsequent experiments. We note that agreement on fluency was lower across both humans and the LLM because modern models rarely make explicit grammatical errors. Therefore, fluency evaluations often shift from objective error-counting to highly subjective stylistic preferences.

\subsection{Structural Metrics}
To complement subjective scores, we apply structural metrics to measure narrative diversity and originality from lexical and semantic perspectives by leveraging embedding-based representations. All of these structural metrics, except readability, are adopted from \citet{ismayilzada2024evaluating} with modifications specific to our setting. Mathematical definitions and implementation details are provided in Appendix~\ref{sec:appendix-metrics}.

\begin{figure*}[b]
    \centering
    \includegraphics[width=\textwidth]{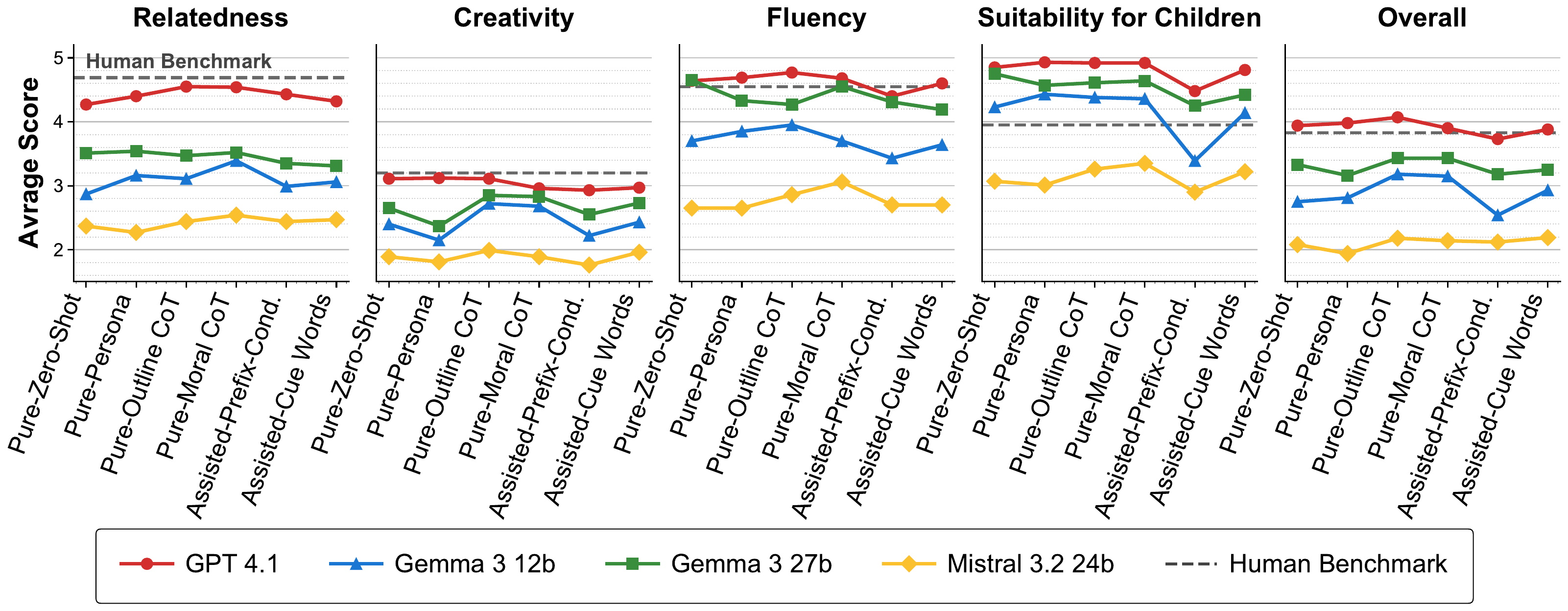}
    \vspace{-1.5em}
    \caption{Average score of LLMs given by our Judge on different metrics for different prompts. Scores of human stories serve as our Human benchmark. For full statistical results (Mean $\pm$ std), refer to Table~\ref{tab:JudgeScores} in Appendix~\ref{sec:detailed-judge}.}
    \label{fig:model_comparison}
\end{figure*}

We quantify \textbf{Diversity} along two complementary dimensions. \textit{Lexical diversity} evaluates surface-level word variation via unique-to-total n-gram ratios ($n \in \{1, \dots, 5\}$). \textit{Semantic diversity} measures the embedding-based cosine distance between the generated story and its human reference text, reflecting the extent to which a model's semantic decompression diverges from human narrative realizations. \textbf{Novelty} measures the semantic uniqueness relative to a broader narrative context~\cite{Maher2010, RuncoJaeger2012}. We compute novelty by comparing a story's embedding against a background pool consisting of all other generated stories for that prompt alongside human stories. Higher values indicate greater deviation from the collective narrative distribution.

\textbf{Surprise}
captures the degree to which a story deviates from expectations as it unfolds~\cite{Boden1991Creative, Maher2010, GraceMaher2014}. We compute surprise as the average semantic distance between consecutive sentences~\cite{Karampiperis2014}, where larger shifts correspond to higher narrative unpredictability. \textbf{Readability} assesses surface-level reading ease in Persian using the Flesch-Dayani metric based on word and sentence length~\cite{dayani2000criteria}. While formal readability metrics correlate poorly with human stylistic preferences~\cite{cachola-etal-2025-evaluating}, we use it as a structural baseline for lexical complexity in combination with other metrics.

\section{Results and Analysis}
\label{sec:results}
We present an extensive evaluation of proprietary and open-weight models across generation regimes. Human-written stories serve as our human benchmark in our experiments.

\textbf{Prompting Strategies Reveal the Decompression Gap:}
\label{sec:results-differetn-prompts}
Figure~\ref{fig:model_comparison} reveals a clear decoupling between surface narrative quality and semantic faithfulness. While GPT-4.1 and the Gemma~3 family approach or even surpass human-written stories in Fluency and Suitability, they exhibit substantially lower Relatedness scores, particularly among smaller open-weight models. This pattern exposes a persistent \emph{decompression gap}: models can generate fluent and well-formed narratives while failing to faithfully instantiate the underlying moral abstraction of the proverb. A similar gap emerges in Creativity, where even the strongest models remain below human performance, highlighting the difficulty of transforming compressed cultural knowledge into engaging narrative form.

Prompting strategy further influences decompression quality. Approaches that explicitly separate semantic interpretation from narrative realization, particularly Moral CoT, consistently achieve higher Relatedness and Overall scores across models. This suggests that extracting and verbalizing the proverb's moral before story generation provides a more effective path to semantic grounding than relying solely on implicit associations or surface-level cues. Notably, Moral CoT in the Pure regime often outperforms strategies in the Surface-Assisted regime, indicating that explicit abstraction may be more beneficial than access to partial narrative context. Detailed results for all model--prompt combinations are reported in Table~\ref{tab:JudgeScores} in Appendix~\ref{sec:detailed-judge}. Examples illustrating the decompression gap are provided in Table~\ref{tab:generated_stories_eval} of Appendix~\ref{sec:appendix-examples}.

\begin{figure*}[t]
    \centering
    \includegraphics[width=\textwidth]{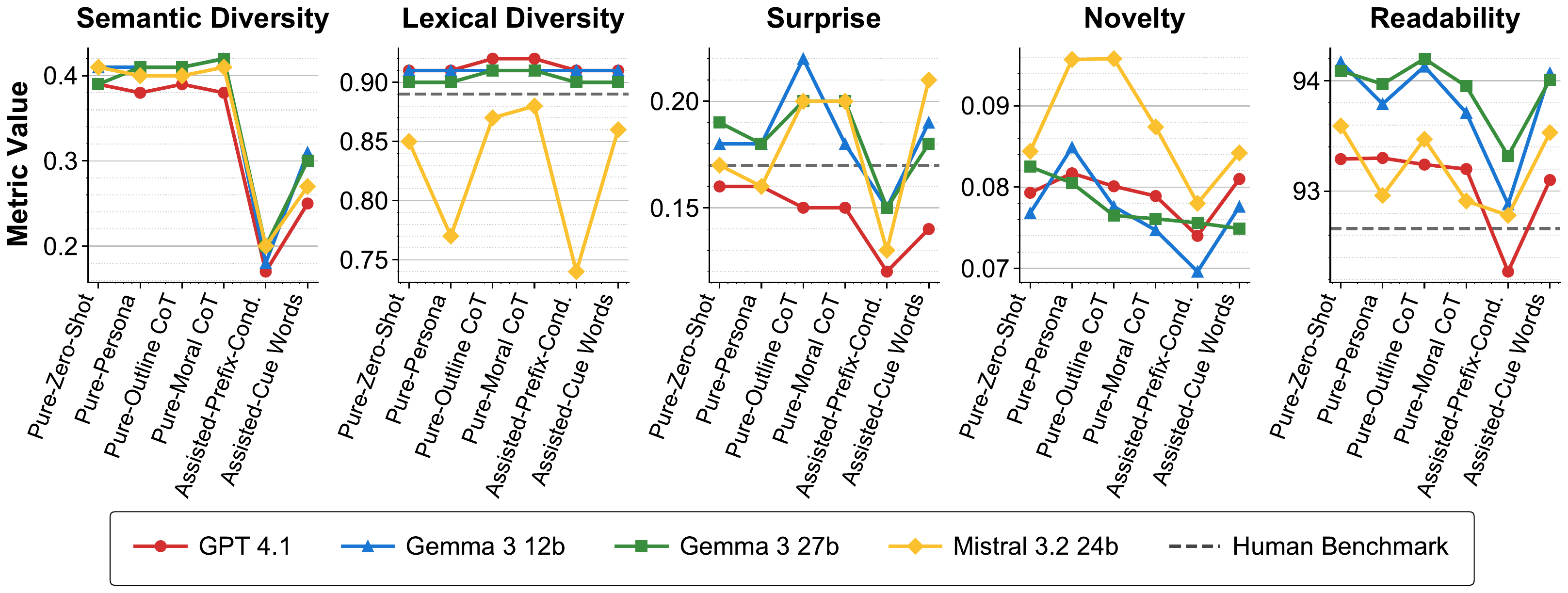}
    \vspace{-1.5em}
    \caption{Performance of different models with different prompts from the lens of structural metrics. Y-axis scaling indicates relative trends of each metric. The average value of each metric for human stories serves as our Human benchmark. For full statistical results (Mean $\pm$ std), refer to Appendix~\ref{sec:detailed-automatic}.}
    \label{fig:model_comparison_automatic}
\end{figure*}

\begin{figure*}[b]
    \centering
    \includegraphics[width=\textwidth]{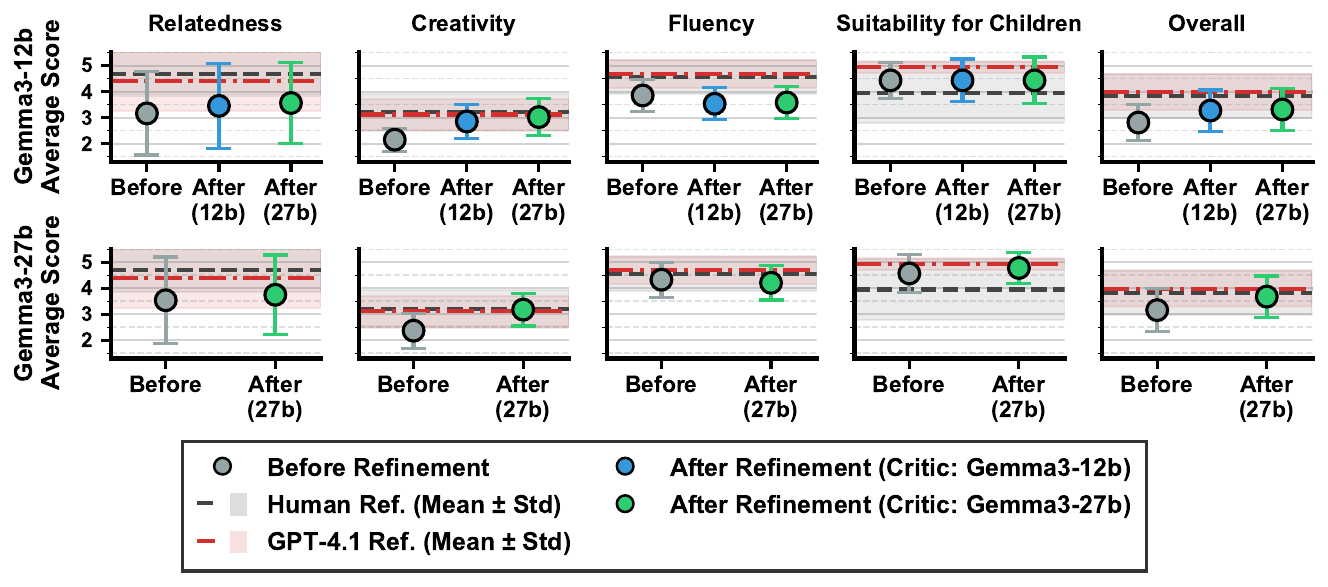}
    \vspace{-2em}
    \caption{Judge scores for Gemma 3 12B (top) and Gemma 3 27B (bottom) before and after feedback-guided refinement (three iterations). Scores of human stories and stories generated with our Persona prompt by GPT 4.1 serve as our Human and GPT 4.1 references, respectively. Full statistical results are provided in Appendix~~\ref{sec:detailed-judge}.}
    \label{fig:iterative}
\end{figure*}

\begin{figure*}[t]
    \centering
    \includegraphics[width=\textwidth]{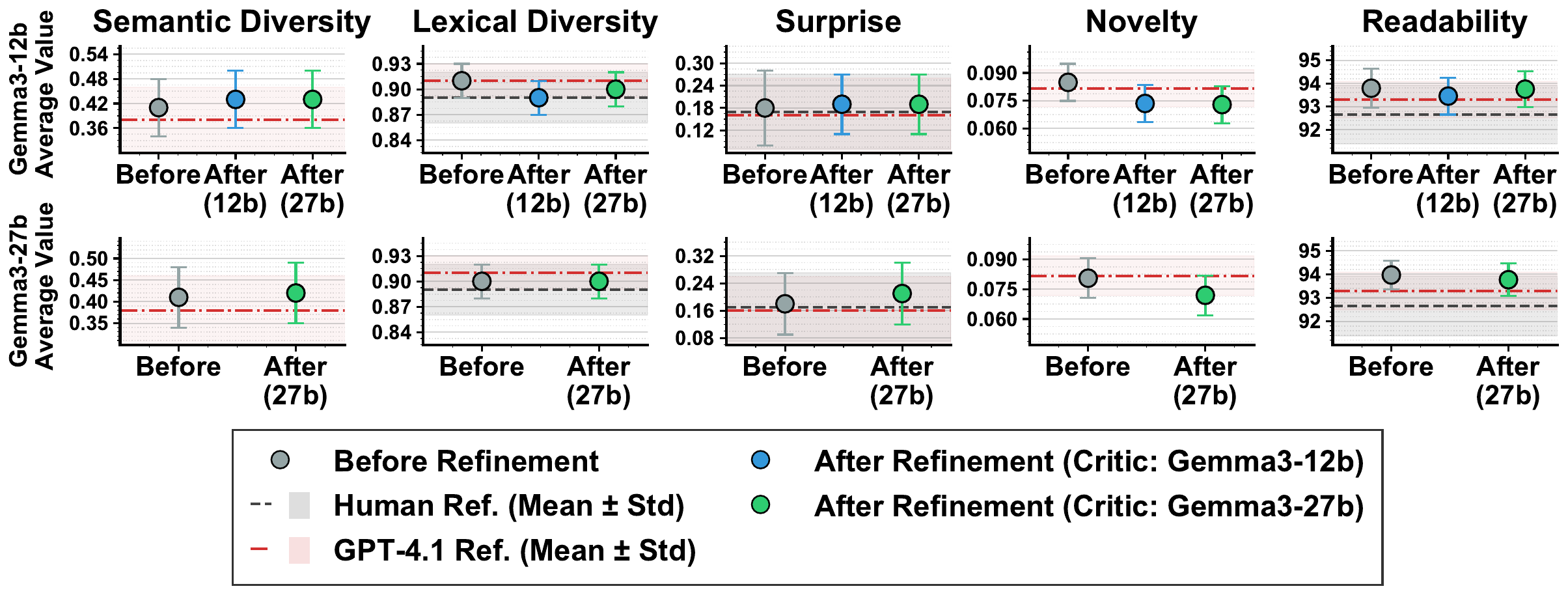}
    \vspace{-2em}
    \caption{Changes in structural metrics values for Gemma 3 12B (top) and Gemma 3 27B (bottom) before and after feedback-guided refinement (three iterations). Scores of human stories and stories generated with our Persona prompt by GPT 4.1 serve as our Human and GPT 4.1 references, respectively. Y-axis scaling indicates relative trends of each metric. Full statistical results are provided in Appendix~\ref{sec:detailed-automatic}.}
    \label{fig:iterative_automatic}
\end{figure*}

\textbf{Analysis of Structural Metrics:}
\label{sec:automatic-detailed}
Figure~\ref{fig:model_comparison_automatic} shows that meaning-level metrics (Semantic Diversity, Surprise, and Novelty) are substantially more sensitive to prompting strategy than surface-level metrics (Lexical Diversity and Readability), which remain relatively stable across regimes. This suggests that modern LLMs have largely saturated basic narrative fluency, while the manner in which they perform semantic decompression varies considerably.

A clear pattern emerges across prompting regimes. The Prefix-Conditioned strategy consistently produces the lowest Semantic Diversity, Surprise, and Novelty scores. By providing the opening sentences of a reference story, this regime constrains the space of plausible narrative realizations and reduces semantic exploration. In contrast, prompts in the Pure regime, particularly the CoT variants, allow models to construct narratives from abstract moral interpretations, resulting in greater semantic dispersion and more diverse decompression trajectories.

Model-level analysis reveals an additional distinction between semantic deviation and narrative quality. Smaller open-weight models tend to generate stories with larger semantic distances from human references and higher Surprise scores than GPT-4.1, indicating less conventional and more unpredictable narrative development. However, greater semantic deviation does not necessarily correspond to better stories. For example, Mistral achieves the highest Novelty scores while simultaneously performing poorly on subjective evaluations, suggesting that its novelty stems from semantic instability rather than meaningful creative exploration. In contrast, GPT-4.1 and Gemma~3~27B exhibit lower novelty but achieve stronger overall performance through more controlled and coherent narrative realizations. These findings highlight the importance of combining subjective and structural evaluations to distinguish grounded creativity from stochastic divergence. Full results are reported in Table~\ref{tab:automatic_metrics_aligned} in Appendix~\ref{sec:detailed-automatic}.

\textbf{Feedback Can Help Decompression:}
\label{sec:results-iterative}
The decompression gap raises a natural question: are abstraction failures intrinsic to open-weight models, or can they be repaired through revision? To investigate this, we apply the Feedback-Guided regime to Gemma~3~12B and~27B under \emph{Self-Correction} and \emph{Cross-Model Critique} (12B writer, 27B critic) settings.

As shown in Figure~\ref{fig:iterative}, iterative refinement consistently improves semantic grounding. For Gemma~3~12B, the stronger 27B critic yields substantially larger gains in Relatedness, Creativity, and Overall score than self-correction, indicating that critique quality is a key factor in repairing decompression failures. Notably, refined Gemma~3~27B surpasses single-pass GPT-4.1 in Creativity while improving Relatedness, Suitability, and Overall quality. These results suggest that many abstraction failures are correctable through targeted feedback and revision.

The gains come with a modest decrease in Fluency, reflecting a surface--semantic trade-off as stories become more tightly aligned with the proverb's moral logic. Structural metrics in Figure~\ref{fig:iterative_automatic} support this interpretation: refinement increases Semantic Diversity and Surprise while slightly reducing Novelty and Lexical Diversity, consistent with convergence toward a more coherent thematic realization. Moreover, the stronger 27B critic incurs smaller reductions in Lexical Diversity and Readability than self-correction, suggesting that higher-capacity critics provide more precise feedback with fewer stylistic compromises. Full results are reported in Appendix~\ref{sec:detailed-results}.

\paragraph{Human Validation of Refinement Improvements:}
To verify that improvements observed by the LLM judge are reflected in human perception, we conducted a blinded pairwise preference study using the Gemma~3~27B self-correction pipeline. Human annotators compared 40 randomly sampled pairs consisting of the initial and refined versions of the same generated story. Based on majority voting, annotators preferred the refined stories in 65\% of cases, providing independent evidence that iterative refinement yields perceptible improvements in narrative quality.

We further examined the relationship between human preferences and changes in LLM-as-a-Judge scores. Most metrics exhibited positive correlations, with Creativity showing the strongest association ($r=0.24$). Although these correlations are not statistically significant given the limited sample size, they indicate broad directional agreement between human judgments and the calibrated judge. Detailed results are reported in Tables~\ref{tab:human_preference} and~\ref{tab:metric_correlations} in Appendix~\ref{sec:metric_correlations}.

\section{Conclusion}

We formalized proverb-conditioned short story generation as a constrained semantic decompression task, where compressed moral abstractions must be transformed into concrete narratives. Using the curated PAND corpus, we evaluated model behavior across multiple prompting regimes with a calibrated judge and complementary structural metrics. Our results reveal a persistent \emph{decompression gap}: current LLMs often achieve strong surface-level fluency while failing to faithfully instantiate the deeper moral and causal structure encoded in proverbs. This finding highlights a fundamental distinction between linguistic competence and semantic grounding in creative generation.

We further show that explicit reasoning and iterative refinement can partially mitigate these failures, suggesting that many decompression errors arise not from a complete lack of knowledge but from difficulties in reliably translating abstract meaning into narrative form. More broadly, our findings indicate that abstraction-to-realization remains a challenging capability even for state-of-the-art models.

In addition to mathematical analysis, future work could internalize the Writer–Critic dynamic via self-play~\cite{chae2025understandingselfplayllmreasoning} to learn refinement policies, replacing explicit prompts. Beyond Persian proverbs, we view semantic decompression as a general framework for studying how LLMs transform compressed knowledge representations, including fables, legal maxims, and moral heuristics, into grounded and contextually appropriate language.

% \clearpage
% \newpage

\section{Limitations}
The primary limitation of this study is the scarcity of high-quality, aligned Persian proverb-story pairs. While our curated dataset is sufficient for evaluation and in-context learning experiments, its scale is inadequate for pre-training or post-training interventions such as supervised fine-tuning or reinforcement learning. As a result, our investigation is necessarily restricted to inference-time prompting strategies, which limits the extent to which models can internalize semantic decompression mechanisms.

A related limitation concerns generalization. Our experiments focus on short narratives and a specific cultural form, proverbs, which may not fully represent other forms of abstract knowledge or longer narrative structures. Although we believe the semantic decompression framework is broadly applicable, its effectiveness for other genres, domains, or languages remains to be empirically validated.

Finally, while we mitigate scalability constraints through a calibrated LLM-as-a-Judge, automated evaluation remains an imperfect proxy for human judgment, particularly for culturally grounded interpretation. Although our calibration shows strong agreement with human annotators, subtle pragmatic or stylistic nuances may still be underrepresented. Future work incorporating larger-scale human evaluation or hybrid human-LLM assessment could further strengthen evaluation reliability.

We hypothesize that access to larger-scale aligned corpora would enable open-weight models to internalize abstraction-to-realization dynamics through training rather than prompting alone. Such training could reduce the surface-semantic trade-off observed in this study and potentially allow open-weight models to achieve state-of-the-art performance in culturally grounded narrative generation.

\section{Ethical Considerations}
Our work primarily addresses story generation for children, a vulnerable audience. While our quantitative and qualitative evaluations indicate that the decompression regimes we investigated produce suitable and non-violent narratives, LLMs remain stochastic by nature. We cannot guarantee 100\% safety against the generation of toxic, biased~\cite{rooein-etal-2025-biased}, or inappropriate content in rare edge cases. Consequently, any real-world deployment of this work would require strict content moderation guardrails and human oversight.

Additionally, our findings are rooted in Persian literary traditions and cultural logic. There is a risk of overgeneralization if these findings are assumed to apply universally to other languages or cultures without empirical validation. Finally, our prompts were optimized for the specific models evaluated in this study; utilizing them in user-facing applications or with untested models without rigorous safety testing could lead to unpredictable or lower-quality outputs.

\bibliography{custom.bib}
\appendix

\section{Data Sources and Example of Dataset}
\label{sec:appendix-data}
This section provides supplementary details about the PAND dataset. Table~\ref{tab:links} lists the source websites along with the corresponding numbers of proverb-story pairs included in the final corpus, and Table~\ref{tab:dataset-example} presents an example instance from the PAND dataset. PAND dataset is intended strictly for non-commercial, academic research purposes to evaluate constrained semantic decompression.

\begin{table*}
  \centering
  \small
  \renewcommand{\arraystretch}{1.4}
  \begin{tabular}{p{0.4\textwidth}r}
    \hline
    \textbf{Link}           & \textbf{Count}\\
    \hline
    \href{https://lelb.net/}{https://lelb.net/} & 73 \\
    \href{https://esanj.ir/}{https://esanj.ir/} & 43 \\
    \href{https://www.kanoon.ir/}{https://www.kanoon.ir/} & 17 \\
    \href{https://setare.com/}
    {https://setare.com/} & 8 \\
    \href{https://roozaneh.net/}{https://roozaneh.net/} & 5 \\
    \href{https://namnak.com/}{https://namnak.com/} & 3 \\
    \href{https://taaghche.com/}{https://taaghche.com/} & 1 \\
    \hline
    Total & 150\\
    \hline
  \end{tabular}
  \caption{\label{tab:links}
    Source websites and counts of proverb-story pairs included in the PAND dataset.
  }
  \renewcommand{\arraystretch}{1.0}
\end{table*}

\begin{table*}[ht]
\centering
\small
% Define a grey color for translations
\definecolor{translationgray}{gray}{0.45}
\newcommand{\transl}[1]{{\color{translationgray}\textit{#1}}}
\begin{tabularx}{\textwidth}{l X} % X column wraps text
\toprule
\textbf{Proverb} & 
\begin{farsi}
بشکنه این دست که نمک نداره.
\end{farsi} \par
\vspace{0.5em}
\transl{(Broken be the hand that has no salt.)} \\
\midrule

\textbf{Meaning} & 
\begin{farsi}
نتیجه عکس گرفتن از خوبی به دیگران.
\end{farsi} \par
\vspace{0.5em}
\transl{(The act of receiving the opposite result of doing good to others.)} \\
\midrule

\textbf{Story} & 
\begin{farsi}
در یک روستای سرسبز و زیبا، یک درخت گردوی تنومند و پربار زندگی می‌کرد. این درخت گردو هر سال بار و میوهٔ خیلی زیادی می‌داد و مردم زیادی از میوه و سایهٔ آن استفاده می‌کردند.

\vspace{0.5em}

این درخت گردو در کنار جاده قرار داشت و به این دلیل، مسافرانی که به روستا می‌آمدند، برای خوردن گردوی خوشمزه و استراحت در سایهٔ درخت، به او پناه می‌آوردند.

\vspace{0.5em}

ولی رفتار مسافران و حتی مردم روستا با این درخت اصلاً خوب نبود. مردم برای اینکه گردوی بیشتری از درخت بچینند و بخورند، همیشه سنگ و چوب به شاخه‌های بالایی درخت گردو پرتاب می‌کردند تا گردوهایی که بر روی شاخه‌های بالاتر هستند، به زمین بریزند.

\vspace{0.5em}

کم‌کم به خاطر این رفتار زشت مردم و ناسپاسی آن‌ها، درخت گردو آسیب زیادی دید و شاخه‌هایش شکست. به این دلیل، دیگر نمی‌توانست مثل قبل، میوه و سایه به مردم بدهد.

\vspace{0.5em}

درخت بیچاره زیر لب با خود گفت: «بشکنه این دست که نمک نداره! همان مردمی که از میوه و سایهٔ من استفاده می‌کردند، به جای تشکر کردن، با سنگ و چوب از من پذیرایی کردند!»
\end{farsi} \par

\vspace{0.5em}

\transl{(In a lush and beautiful village, there lived a sturdy and fruitful walnut tree. Every year, it gave plenty of fruit, and many people enjoyed its shade. Since it was by the road, travelers sought refuge under it. However, the people were ungrateful. They threw stones and sticks to knock down the walnuts, breaking its branches. Eventually, the tree could no longer provide fruit or shade. The poor tree whispered to itself: ``Broken be the hand that has no salt! The same people who enjoyed my fruit and shade treated me with stones instead of thanks!'' )} \\
\bottomrule
\end{tabularx}
\caption{\label{tab:dataset-example} An example from the PAND Dataset. Literal English translations are provided in parentheses in gray below the original Persian text. These translations are not included in the prompts presented to the model.}
\end{table*}

% \section{Models}
% \label{sec:appendix-models}

\section{Experimental Details}
\label{sec:appendix-experiment}
To automatically generate Persian stories, we utilize a combination of proprietary and open-weight LLMs with strong Persian support.  We specifically focus on smaller, efficient models to evaluate whether they can perform comparably to larger closed-source systems. Our final set of models comprises GPT-4.1~\citep{openai2024gpt4technicalreport}, Gemma~3 (12B and 27B)~\citep{gemmateam2025gemma3technicalreport}, and Mistral Small~3.2~24B~\citep{mistralModel}. We additionally evaluated several other models for story generation, including Gemma3 4B ~\citep{gemmateam2025gemma3technicalreport}, Qwen3 (4B and 14B) \citep{yang2025qwen3technicalreport}, DeepSeek-V3.2 \citep{deepseekai2025deepseekv32pushingfrontieropen}, GPT-OSS-20B \citep{openai2025gptoss120bgptoss20bmodel}, and Mistral Small Creative \citep{mistralModelCreative}. However, their performance was not reportable, as they frequently produced nonsensical outputs, non-Persian characters, or content unrelated to the target proverb. We access all models via the OpenRouter API\footnote{\url{https://openrouter.ai}} using the default temperature of 1 to encourage diverse and creative responses. All other parameters are also set to their default values. The total cost of running the experiments amounts to approximately \$100 USD. It is notable to mention that all models were accessed and employed in strict accordance with their respective licenses and Terms of Service, as well as their acceptable and intended use policies.

\section{Details of Feedback-Guided Semantic Decompression Experiment}
\label{sec:appendix-refine-detail}
To ensure the reproducibility of the feedback-guided regime experiments, we provide an algorithmic overview in Algorithm~\ref{alg:story_loop} and the specific configuration details below.

\paragraph{Model Configurations}
We utilized the Gemma 3 family of open-weight models. The Self-Correction setup involved a single model instance (either Gemma 3 12B or 27B) performing all roles. The Cross-Model Critique setup utilized Gemma 3 12B as the Writer and Editor, with Gemma 3 27B acting as the Critic.

\begin{algorithm}
\caption{Iterative Story Generation}
\label{alg:story_loop}
\begin{algorithmic}[1]
\State \textbf{Input:} Proverb $P$, Iterations $K$
\State \textbf{Output:} Final Story $S$
\State $S \leftarrow \text{Gen}(P)$
\For{$i \leftarrow 1$ \textbf{to} $K$}
    \State $Sc, Exp, Rec \leftarrow \text{Critic}(S, P)$
    \State $S \leftarrow \text{Editor}(S, Sc, Exp, Rec)$
\EndFor
\State \Return $S$
\end{algorithmic}
\end{algorithm}

\paragraph{Refinement Loop}
The process was executed for a fixed duration of three iterations. To ensure a strictly controlled comparison between the self-correction and cross-model settings, we generated a single set of initial drafts (Iteration 0) using the Writer persona and froze these drafts. Both the Self-Correction and Cross-Model pipelines then operated on this identical starting set. This ensures that any observed performance differences are attributable solely to the quality of the critique and editing, rather than random variations in the initial generation. To compare the effect of this refinement loop, we evaluate the initial draft (iteration 0, before entering the refinement loop) and the final draft (iteration 3, after finishing the refinement loop) utilizing our LLM-as-a-Judge and automatic scores.

\section{Calibration and Selection of LLM-as-a-Judge}
\label{sec:appendix-judge}
 To ensure our automatic evaluation acts as a reliable proxy for human judgment, particularly regarding the cultural nuances of Persian proverbs, we conducted an extensive calibration study. This section details the candidate models, agreement methodologies, and quantitative results that led to the selection of Gemini 2.5 Pro as our evaluator.

\subsection{Candidate Models}
We evaluated a diverse set of LLMs to serve as the judge. Our selection criteria balanced SOTA performance with the potential accessibility of open-weight models.

\paragraph{Proprietary SOTA Models}
To maximize the likelihood of capturing deep semantic and cultural alignment, we tested the strongest available closed-source models, including GPT-5 \cite{openai2025gpt5}, GPT-4o \cite{openai2024gpt4ocard}, Claude Sonnet 4 \cite{anthropic2025claude4}, and Gemini 2.5 Pro ~\cite{comanici2025gemini25pushingfrontier}.

\paragraph{Open-Weight Models}
We also investigated open-weight models to determine whether a cost-effective, locally hosted judge could achieve human parity, including DeepSeek R1 \cite{deepseekai2025deepseekr1incentivizingreasoningcapability} and DeepSeek v3 \cite{deepseekai2025deepseekv3technicalreport}, Gemma 3 4B \citep{gemmateam2025gemma3technicalreport}, and Mistral Small 3.2 24B \citep{mistralModel}.

\subsection{Methodology and Aggregation Strategies}
We utilized the dataset described in Section \ref{sec:subjective_evaluation}, consisting of 40 human-written and 40 GPT-4.1-generated stories using the Prefix-Conditioned prompt (80 total), annotated by three human experts. Each candidate LLM was prompted to rate these stories on the same Likert scale used by humans and against the same qualitative metrics (Relatedness, Creativity, Fluency, Suitability, and Overall Quality).
To rigorously quantify the alignment between the LLM judge and human perception, we calculated four agreement metrics: ICC(2,k), ICC(3,k), Krippendorff’s Alpha, and Exact Match accuracy. Furthermore, acknowledging that human agreement is not monolithic, we computed these metrics using four distinct aggregation strategies to verify robustness:
\begin{enumerate}
\item \textbf{LLM vs. Median:} Agreement between the LLM and the median human score, mitigating outlier influence.
\item \textbf{LLM vs. Mean:} Agreement between the LLM and the average human score.
\item \textbf{Pooled Consensus (All):} Treating the LLM as a fourth independent annotator alongside the three humans to calculate group inter-rater reliability.
\end{enumerate}

\subsection{Results and Selection}
The comparative results of these experiments are presented in Table \ref{tab:combined-agreement}.

\paragraph{SOTA vs. Open-Weight Gap} A clear dichotomy emerged between proprietary and open-weight models. While open-weight models (e.g., Mistral, Gemma, DeepSeek) offer efficiency, they consistently showed poor correlation with human judges on this task. This suggests that current open-weight models generally lack the cultural depth or instruction-following precision required for Persian semantic decompression evaluation.

\paragraph{Selection of Gemini 2.5 Pro} Among the SOTA models, the competition was tight. GPT-5 and GPT-4o demonstrated strong alignment, performing nearly on par with the leader in several categories. However, Gemini 2.5 Pro emerged as the most robust model overall, achieving the highest agreement scores in 3 out of the 5 evaluation criteria (specifically Fluency, Suitability for Children, and Overall) within the Pooled Consensus setting. Furthermore, it maintained highly competitive performance in Relatedness and Creativity. Given its superior alignment on pragmatic constraints (Suitability) and overall consistency, we adopted Gemini 2.5 Pro as the fixed evaluator for all subsequent  experiments.

\begin{table*}[t]
\centering
\scriptsize
\setlength{\tabcolsep}{2.5pt} % Reduce padding to fit all columns
\begin{tabular}{lll | ccc | ccc | ccc | ccc | ccc}
\toprule
\multirow{2}{*}{\textbf{Type}} & \multirow{2}{*}{\textbf{Model}} & \multirow{2}{*}{\textbf{Measure}} & \multicolumn{3}{c}{\textbf{Relatedness}} & \multicolumn{3}{c}{\textbf{Creativity}} & \multicolumn{3}{c}{\textbf{Fluency}} & \multicolumn{3}{c}{\textbf{Suitability for Ch.}} & \multicolumn{3}{c}{\textbf{Overall}} \\
\cmidrule(lr){4-6} \cmidrule(lr){7-9} \cmidrule(lr){10-12} \cmidrule(lr){13-15} \cmidrule(lr){16-18}
 & & & Mean & Med & All & Mean & Med & All & Mean & Med & All & Mean & Med & All & Mean & Med & All \\
\midrule

% CLOSED SOURCE MODELS
\multirow{16}{*}{\rotatebox{90}{\textbf{Closed-Source}}} 
& \multirow{4}{*}{\textbf{GPT-4o}} 
  & ICC2k & 0.67 & 0.69 & 0.84 & 0.69 & 0.66 & 0.82 & 0.26 & 0.30 & 0.44 & 0.22 & 0.27 & 0.60 & 0.55 & 0.51 & 0.77 \\
& & ICC3k & 0.71 & 0.72 & 0.86 & \textbf{0.69} & \textbf{0.66} & \textbf{0.83} & \textbf{0.45} & \textbf{0.43} & 0.52 & 0.42 & 0.45 & 0.72 & \textbf{0.72} & \textbf{0.69} & \textbf{0.83} \\
& & Krip's $\alpha$ & 0.48 & 0.51 & 0.57 & 0.53 & 0.49 & 0.54 & -0.12 & 0.00 & 0.12 & -0.18 & -0.09 & 0.22 & 0.25 & 0.21 & 0.43 \\
& & Match & 14 & 41 & 13 & 20 & 50 & 16 & 12 & 43 & 11 & 3 & 23 & 3 & 5 & 20 & 4 \\
\cmidrule{2-18}

& \multirow{4}{*}{\textbf{GPT-5}} 
  & ICC2k & 0.78 & 0.79 & 0.87 & 0.41 & 0.41 & 0.76 & 0.31 & 0.30 & 0.47 & 0.34 & 0.40 & 0.65 & 0.68 & 0.65 & 0.81 \\
& & ICC3k & \textbf{0.81} & \textbf{0.81} & \textbf{0.89} & 0.48 & 0.47 & 0.78 & 0.31 & 0.30 & 0.48 & 0.45 & 0.49 & 0.71 & 0.68 & 0.66 & 0.82 \\
& & Krip's $\alpha$ & 0.62 & 0.64 & 0.62 & 0.18 & 0.19 & 0.43 & 0.18 & 0.18 & 0.17 & 0.08 & 0.15 & 0.29 & 0.51 & 0.48 & 0.52 \\
& & Match & 16 & 44 & 15 & 8 & 30 & 6 & 12 & 40 & 12 & 5 & 24 & 3 & 24 & 49 & 20 \\
\cmidrule{2-18}

& \multirow{4}{*}{\textbf{Sonnet 4}} 
  & ICC2k & 0.69 & 0.68 & 0.85 & 0.64 & 0.56 & 0.81 & 0.27 & 0.22 & 0.45 & 0.39 & 0.45 & 0.67 & 0.63 & 0.59 & 0.80 \\
& & ICC3k & 0.71 & 0.69 & 0.86 & 0.64 & 0.56 & 0.81 & 0.32 & 0.26 & 0.48 & 0.43 & 0.48 & 0.70 & 0.63 & 0.60 & 0.81 \\
& & Krip's $\alpha$ & 0.52 & 0.51 & 0.58 & 0.46 & 0.39 & 0.52 & 0.06 & 0.03 & 0.14 & 0.19 & 0.26 & 0.33 & 0.45 & 0.42 & 0.50 \\
& & Match & 16 & 41 & 15 & 18 & 48 & 13 & 12 & 40 & 11 & 8 & 30 & 7 & 23 & 48 & 20 \\
\cmidrule{2-18}

& \multirow{4}{*}{\shortstack{\textbf{Gemini}\\\textbf{2.5 Pro}}} 
  & ICC2k & 0.69 & 0.68 & 0.85 & 0.40 & 0.37 & 0.73 & 0.40 & 0.36 & 0.51 & 0.63 & 0.65 & 0.75 & 0.73 & 0.66 & 0.82 \\
& & ICC3k & 0.74 & 0.72 & 0.87 & 0.55 & 0.51 & 0.79 & 0.43 & 0.38 & \textbf{0.53} & \textbf{0.68} & \textbf{0.69} & \textbf{0.78} & \textbf{0.72} & 0.66 & \textbf{0.83} \\
& & Krip's $\alpha$ & 0.51 & 0.50 & 0.57 & 0.09 & 0.08 & 0.37 & 0.22 & 0.20 & 0.19 & 0.42 & 0.46 & 0.42 & 0.57 & 0.49 & 0.54 \\
& & Match & 16 & 37 & 16 & 8 & 24 & 6 & 13 & 40 & 12 & 13 & 36 & 13 & 21 & 45 & 17 \\
\midrule

% OPEN WEIGHT MODELS
\multirow{16}{*}{\rotatebox{90}{\textbf{Open-Weight}}} 
& \multirow{4}{*}{\shortstack{\textbf{DeepSeek}\\\textbf{R1}}} 
  & ICC2k & 0.47 & 0.48 & 0.80 & 0.34 & 0.26 & 0.75 & 0.27 & 0.32 & 0.45 & 0.17 & 0.23 & 0.59 & 0.39 & 0.37 & 0.74 \\
& & ICC3k & 0.59 & 0.57 & 0.84 & 0.34 & 0.26 & 0.76 & 0.31 & 0.35 & 0.48 & 0.31 & 0.39 & 0.70 & 0.48 & 0.47 & 0.77 \\
& & Krip's $\alpha$ & 0.19 & 0.23 & 0.48 & 0.19 & 0.14 & 0.43 & 0.08 & 0.14 & 0.14 & -0.22 & -0.12 & 0.21 & 0.14 & 0.11 & 0.39 \\
& & Match & 16 & 41 & 14 & 13 & 37 & 9 & 11 & 42 & 11 & 4 & 22 & 4 & 17 & 33 & 14 \\
\cmidrule{2-18}

& \multirow{4}{*}{\shortstack{\textbf{DeepSeek}\\\textbf{V3}}} 
  & ICC2k & 0.37 & 0.39 & 0.77 & 0.36 & 0.35 & 0.77 & 0.06 & 0.04 & 0.36 & 0.29 & 0.33 & 0.63 & 0.29 & 0.26 & 0.67 \\
& & ICC3k & 0.53 & 0.52 & 0.84 & 0.39 & 0.37 & 0.78 & 0.16 & 0.07 & 0.45 & 0.50 & 0.51 & 0.73 & 0.57 & 0.53 & 0.80 \\
& & Krip's $\alpha$ & 0.06 & 0.11 & 0.44 & 0.19 & 0.17 & 0.45 & -0.36 & -0.25 & 0.06 & -0.08 & 0.00 & 0.25 & -0.16 & -0.19 & 0.28 \\
& & Match & 15 & 37 & 14 & 19 & 49 & 15 & 12 & 37 & 12 & 5 & 29 & 5 & 5 & 13 & 4 \\
\cmidrule{2-18}

& \multirow{4}{*}{\textbf{Gemma 3}} 
  & ICC2k & 0.46 & 0.52 & 0.78 & 0.09 & 0.09 & 0.64 & -0.03 & -0.02 & 0.13 & 0.17 & 0.23 & 0.69 & -0.05 & -0.01 & 0.62 \\
& & ICC3k & 0.46 & 0.52 & 0.79 & 0.11 & 0.11 & 0.68 & -0.04 & -0.02 & 0.15 & 0.17 & 0.23 & 0.61 & -0.05 & -0.01 & 0.64 \\
& & Krip's $\alpha$ & 0.29 & 0.34 & 0.46 & -0.05 & -0.04 & 0.29 & -0.12 & -0.11 & 0.00 & 0.08 & 0.13 & 0.26 & -0.07 & -0.04 & 0.27 \\
& & Match & 9 & 32 & 9 & 16 & 37 & 11 & 8 & 26 & 7 & 11 & 26 & 9 & 20 & 33 & 16 \\
\cmidrule{2-18}

& \multirow{4}{*}{\textbf{Mistral 3.2}} 
  & ICC2k & 0.45 & 0.45 & 0.80 & 0.65 & 0.61 & 0.81 & 0.24 & -0.05 & 0.44 & 0.47 & 0.50 & 0.71 & 0.52 & 0.49 & 0.78 \\
& & ICC3k & 0.45 & 0.45 & 0.81 & 0.65 & 0.61 & 0.82 & 0.25 & -0.05 & 0.46 & 0.49 & 0.51 & 0.73 & 0.53 & 0.51 & 0.79 \\
& & Krip's $\alpha$ & 0.29 & 0.29 & 0.50 & 0.48 & 0.44 & 0.52 & 0.11 & -0.03 & 0.15 & 0.29 & 0.33 & 0.37 & 0.34 & 0.31 & 0.46 \\
& & Match & 12 & 41 & 10 & 17 & 44 & 13 & 13 & 35 & 12 & 16 & 42 & 12 & 24 & 48 & 19 \\

\bottomrule
\end{tabular}
\caption{Inter-annotator agreement between human annotators and both closed-source and open-source LLMs across various metrics. \textbf{Bold} indicates the highest ICC(3,k) agreement score for each metric across all models.}
\label{tab:combined-agreement}
\end{table*}

\section{Detailed Results}
\label{sec:detailed-results}

This section provides complete numerical results for all experiments reported in the main paper, including both LLM-as-a-judge evaluations and structural metrics, across our different proverb decompression strategies.

\subsection{LLM-as-a-Judge}
\label{sec:detailed-judge}
Table \ref{tab:JudgeScores} presents judge scores (Mean $\pm$ Standard Deviation) for all evaluated models on non-iterative prompts from the pure and surface-assisted regimes. Table \ref{tab:iterative-results} provides the corresponding scores for experiments under the iterative refinement framework of the feedback-guided regime. All scores were generated by our judge (Gemini 2.5 Pro) using the 5-point Likert scale described in Section \ref{sec:subjective_evaluation}.

\subsection{Structural Metrics}
\label{sec:detailed-automatic}
Table \ref{tab:automatic_metrics_aligned} provides structural metrics for non-iterative prompts from the pure and surface-assisted regimes. Table \ref{tab:iterative-results-automatic} reports the corresponding metrics for experiments under the iterative refinement framework of the feedback-guided regime.

% ############################

\begin{table*}[t]
\centering
\small
\renewcommand{\arraystretch}{1.1}

\resizebox{\textwidth}{!}{%
\begin{tabular}{l l l c c c c c}
    \toprule
    \multicolumn{3}{c}{\textbf{Story Details}} & \multicolumn{5}{c}{\textbf{Judge Scores}} \\
    \cmidrule(r){1-3} \cmidrule(l){4-8}
    
    \textbf{Decompression Regime} & \textbf{Prompt} & \textbf{Writer/Model} & 
    \textbf{Relatedness} & \textbf{Creativity} & \textbf{Fluency} & 
    \textbf{\makecell{Suitability\\for Children}} & \textbf{Overall} \\ 
    \midrule

    -- & -- & \textbf{Human} & $4.69 \pm 0.84$ & $3.20 \pm 0.72$ & $4.55 \pm 0.66$ & $3.95 \pm 1.17$ & $3.83 \pm 0.86$ \\ 
    \midrule

    \multirow{16}{*}{\textbf{Pure}} 
    & \multirow{4}{*}{\textbf{Zero-Shot}} 
    & GPT-4.1 & $4.27 \pm 1.31$ & $3.11 \pm 0.69$ & $4.64 \pm 0.56$ & $4.85 \pm 0.49$ & $3.94 \pm 0.82$ \\
    & & Gemma-3-12B & $2.87 \pm 1.63$ & $2.40 \pm 0.61$ & $3.70 \pm 0.83$ & $4.23 \pm 1.01$ & $2.75 \pm 0.93$ \\
    & & Gemma-3-27B & $3.51 \pm 1.55$ & $2.65 \pm 0.69$ & $4.65 \pm 0.56$ & $4.75 \pm 0.61$ & $3.33 \pm 0.96$ \\
    & & Mistral-3.2-24B & $2.37 \pm 1.54$ & $1.89 \pm 0.59$ & $2.65 \pm 0.79$ & $3.07 \pm 0.97$ & $2.08 \pm 0.78$ \\
    \cmidrule{2-8}
    
    & \multirow{4}{*}{\textbf{Persona}} 
    & GPT-4.1 & $4.40 \pm 1.16$ & $3.12 \pm 0.59$ & $4.69 \pm 0.54$ & $4.93 \pm 0.23$ & $3.98 \pm 0.68$ \\
    & & Gemma-3-12B & $3.16 \pm 1.60$ & $2.15 \pm 0.44$ & $3.85 \pm 0.63$ & $4.43 \pm 0.70$ & $2.81 \pm 0.69$ \\
    & & Gemma-3-27B & $3.54 \pm 1.66$ & $2.37 \pm 0.51$ & $4.33 \pm 0.67$ & $4.57 \pm 0.74$ & $3.16 \pm 0.83$ \\
    & & Mistral-3.2-24B & $2.27 \pm 1.51$ & $1.81 \pm 0.63$ & $2.65 \pm 0.86$ & $3.01 \pm 1.10$ & $1.94 \pm 0.82$ \\
    \cmidrule{2-8}

    & \multirow{4}{*}{\makecell[l]{\textbf{Outline CoT}}} 
    & GPT-4.1 & $4.55 \pm 1.05$ & $3.11 \pm 0.61$ & $4.77 \pm 0.47$ & $4.92 \pm 0.35$ & $4.07 \pm 0.67$ \\
    & & Gemma-3-12B & $3.11 \pm 1.60$ & $2.72 \pm 0.72$ & $3.95 \pm 0.86$ & $4.38 \pm 0.86$ & $3.18 \pm 1.00$ \\
    & & Gemma-3-27B & $3.47 \pm 1.59$ & $2.85 \pm 0.68$ & $4.27 \pm 0.77$ & $4.61 \pm 0.76$ & $3.43 \pm 0.97$ \\ 
    & & Mistral-3.2-24B & $2.44 \pm 1.60$ & $1.99 \pm 0.55$ & $2.86 \pm 0.80$ & $3.26 \pm 0.94$ & $2.18 \pm 0.83$ \\
    \cmidrule{2-8}

    & \multirow{4}{*}{\makecell[l]{\textbf{Moral CoT}}} 
    & GPT-4.1 & $4.54 \pm 1.02$ & $2.96 \pm 0.65$ & $4.68 \pm 0.66$ & $4.92 \pm 0.29$ & $3.90 \pm 0.64$ \\
    & & Gemma-3-12B & $3.39 \pm 1.61$ & $2.68 \pm 0.59$ & $3.70 \pm 0.82$ & $4.36 \pm 0.78$ & $3.15 \pm 0.90$ \\
    & & Gemma-3-27B & $3.52 \pm 1.57$ & $2.83 \pm 0.73$ & $4.55 \pm 0.65$ & $4.64 \pm 0.71$ & $3.43 \pm 0.92$ \\ 
    & & Mistral-3.2-24B & $2.54 \pm 1.65$ & $1.89 \pm 0.53$ & $3.06 \pm 0.84$ & $3.35 \pm 0.99$ & $2.14 \pm 0.81$ \\
    \midrule

    \multirow{8}{*}{\textbf{Surface-Assisted}} 
    & \multirow{4}{*}{\makecell[l]{\textbf{Prefix-Conditioned}}} 
    & GPT-4.1 & $4.43 \pm 1.18$ & $2.93 \pm 0.63$ & $4.40 \pm 0.64$ & $4.48 \pm 0.86$ & $3.73 \pm 0.80$ \\
    & & Gemma-3-12B & $2.99 \pm 1.70$ & $2.22 \pm 0.60$ & $3.43 \pm 0.86$ & $3.39 \pm 1.06$ & $2.54 \pm 0.88$ \\
    & & Gemma-3-27B & $3.35 \pm 1.57$ & $2.55 \pm 0.56$ & $4.31 \pm 0.72$ & $4.25 \pm 0.89$ & $3.18 \pm 0.86$ \\ 
    & & Mistral-3.2-24B & $2.44 \pm 1.56$ & $1.76 \pm 0.62$ & $2.70 \pm 1.02$ & $2.90 \pm 1.22$ & $2.12 \pm 0.90$ \\
    \cmidrule{2-8}

    & \multirow{4}{*}{\makecell[l]{\textbf{Cue-Words}}} 
    & GPT-4.1 & $4.32 \pm 1.12$ & $2.97 \pm 0.67$ & $4.60 \pm 0.54$ & $4.81 \pm 0.49$ & $3.88 \pm 0.74$ \\
    & & Gemma-3-12B & $3.06 \pm 1.60$ & $2.43 \pm 0.57$ & $3.64 \pm 0.83$ & $4.14 \pm 1.01$ & $2.93 \pm 0.91$ \\
    & & Gemma-3-27B & $3.31 \pm 1.62$ & $2.73 \pm 0.73$ & $4.19 \pm 0.72$ & $4.42 \pm 0.80$ & $3.25 \pm 1.02$ \\
    & & Mistral-3.2-24B & $2.47 \pm 1.55$ & $1.96 \pm 0.54$ & $2.70 \pm 0.72$ & $3.22 \pm 0.99$ & $2.19 \pm 0.72$ \\
    
    \bottomrule
\end{tabular}%
}
\caption{LLM-as-a-judge scores (Gemini 2.5 Pro) for human and model-generated stories under different prompting strategies.}
\label{tab:JudgeScores}
\end{table*}

\begin{table*}[t]
\centering
\small
\renewcommand{\arraystretch}{1.1}

\resizebox{\textwidth}{!}{%
\begin{tabular}{l l c c c c c c}
    \toprule
    \multicolumn{3}{c}{\textbf{Story Details}} & 
    \multicolumn{5}{c}{\textbf{Structural Metrics}} \\
    \cmidrule(r){1-3} \cmidrule(l){4-8}

    \textbf{Decompression Regime} & \textbf{Prompt} & \textbf{Writer/Model} &
    \textbf{\makecell{Semantic\\Diversity}} &
    \textbf{\makecell{Lexical\\Diversity}} &
    \textbf{Surprise} &
    \textbf{Novelty} &
    \textbf{Readability} \\
    \midrule

    -- & -- & \textbf{Human} & -- & $0.89 \pm 0.03$ & $0.17 \pm 0.10$ & -- & $92.66 \pm 1.27$ \\
    \midrule

    \multirow{16}{*}{\textbf{Pure}}
      & \multirow{4}{*}{\textbf{Zero-Shot}}
      & GPT-4.1           & $0.39 \pm 0.07$ & $0.91 \pm 0.02$ & $0.16 \pm 0.08$ & $7.93E\text{-}2 \pm 0.01$ & $93.29 \pm 0.75$ \\
      & & Gemma-3-12B       & $0.41 \pm 0.07$ & $0.91 \pm 0.03$ & $0.18 \pm 0.10$ & $7.68E\text{-}2 \pm 0.02$ & $94.17 \pm 0.77$ \\
      & & Gemma-3-27B       & $0.39 \pm 0.07$ & $0.90 \pm 0.03$ & $0.19 \pm 0.10$ & $8.25E\text{-}2 \pm 0.01$ & $94.09 \pm 0.63$ \\
      & & Mistral-3.2-24B   & $0.41 \pm 0.07$ & $0.85 \pm 0.16$ & $0.17 \pm 0.10$ & $8.44E\text{-}2 \pm 0.02$ & $93.59 \pm 0.94$ \\
    \cmidrule{2-8}

      & \multirow{4}{*}{\textbf{Persona}}
      & GPT-4.1           & $0.38 \pm 0.08$ & $0.91 \pm 0.02$ & $0.16 \pm 0.10$ & $8.17E\text{-}2 \pm 0.01$ & $93.30 \pm 0.80$ \\
      & & Gemma-3-12B       & $0.41 \pm 0.07$ & $0.91 \pm 0.02$ & $0.18 \pm 0.10$ & $8.49E\text{-}2 \pm 0.01$ & $93.79 \pm 0.85$ \\
      & & Gemma-3-27B       & $0.41 \pm 0.07$ & $0.90 \pm 0.02$ & $0.18 \pm 0.09$ & $8.05E\text{-}2 \pm 0.01$ & $93.97 \pm 0.61$ \\
      & & Mistral-3.2-24B   & $0.40 \pm 0.07$ & $0.77 \pm 0.20$ & $0.16 \pm 0.10$ & $9.57E\text{-}2 \pm 0.01$ & $92.96 \pm 0.80$ \\
    \cmidrule{2-8}

      & \multirow{4}{*}{\makecell[l]{\textbf{Outline CoT}}}
      & GPT-4.1           & $0.39 \pm 0.08$ & $0.92 \pm 0.02$ & $0.15 \pm 0.09$ & $8.01E\text{-}2 \pm 0.01$ & $93.24 \pm 0.84$ \\
      & & Gemma-3-12B       & $0.41 \pm 0.07$ & $0.91 \pm 0.02$ & $0.22 \pm 0.11$ & $7.76E\text{-}2 \pm 0.01$ & $94.13 \pm 0.77$ \\
      & & Gemma-3-27B       & $0.41 \pm 0.07$ & $0.91 \pm 0.02$ & $0.20 \pm 0.11$ & $7.65E\text{-}2 \pm 0.01$ & $94.20 \pm 0.76$ \\
      & & Mistral-3.2-24B   & $0.40 \pm 0.07$ & $0.87 \pm 0.05$ & $0.20 \pm 0.14$ & $9.58E\text{-}2 \pm 0.02$ & $93.47 \pm 0.81$ \\
    \cmidrule{2-8}

      & \multirow{4}{*}{\makecell[l]{\textbf{Moral CoT}}}
      & GPT-4.1           & $0.38 \pm 0.08$ & $0.92 \pm 0.02$ & $0.15 \pm 0.10$ & $7.89E\text{-}2 \pm 0.01$ & $93.20 \pm 0.77$ \\
      & & Gemma-3-12B       & $0.42 \pm 0.06$ & $0.91 \pm 0.02$ & $0.18 \pm 0.09$ & $7.47E\text{-}2 \pm 0.01$ & $93.71 \pm 0.76$ \\
      & & Gemma-3-27B       & $0.42 \pm 0.07$ & $0.91 \pm 0.03$ & $0.20 \pm 0.11$ & $7.61E\text{-}2 \pm 0.01$ & $93.95 \pm 0.81$ \\
      & & Mistral-3.2-24B   & $0.41 \pm 0.07$ & $0.88 \pm 0.05$ & $0.20 \pm 0.13$ & $8.74E\text{-}2 \pm 0.01$ & $92.91 \pm 0.83$ \\
    \midrule

    \multirow{8}{*}{\textbf{Surface-Assisted}}
      & \multirow{4}{*}{\makecell[l]{\textbf{Prefix-Conditioned}}}
      & GPT-4.1           & $0.17 \pm 0.07$ & $0.91 \pm 0.02$ & $0.12 \pm 0.08$ & $7.40E\text{-}2 \pm 0.01$ & $92.27 \pm 1.13$ \\
      & & Gemma-3-12B       & $0.18 \pm 0.07$ & $0.91 \pm 0.02$ & $0.15 \pm 0.10$ & $6.96E\text{-}2 \pm 0.01$ & $92.88 \pm 1.25$ \\
      & & Gemma-3-27B       & $0.20 \pm 0.08$ & $0.90 \pm 0.03$ & $0.15 \pm 0.10$ & $7.56E\text{-}2 \pm 0.01$ & $93.32 \pm 0.92$ \\
      & & Mistral-3.2-24B   & $0.20 \pm 0.08$ & $0.74 \pm 0.31$ & $0.13 \pm 0.11$ & $7.80E\text{-}2 \pm 0.02$ & $92.78 \pm 1.12$ \\
    \cmidrule{2-8}

      & \multirow{4}{*}{\makecell[l]{\textbf{Cue-Words}}}
      & GPT-4.1           & $0.25 \pm 0.08$ & $0.91 \pm 0.02$ & $0.14 \pm 0.08$ & $8.10E\text{-}2 \pm 0.01$ & $93.10 \pm 0.80$ \\
      & & Gemma-3-12B       & $0.31 \pm 0.07$ & $0.91 \pm 0.02$ & $0.19 \pm 0.09$ & $7.76E\text{-}2 \pm 0.01$ & $94.07 \pm 0.71$ \\
      & & Gemma-3-27B       & $0.30 \pm 0.07$ & $0.90 \pm 0.02$ & $0.18 \pm 0.09$ & $7.49E\text{-}2 \pm 0.01$ & $94.01 \pm 0.74$ \\
      & & Mistral-3.2-24B   & $0.27 \pm 0.06$ & $0.86 \pm 0.09$ & $0.21 \pm 0.12$ & $8.42E\text{-}2 \pm 0.01$ & $93.53 \pm 0.82$ \\
    \bottomrule
\end{tabular}%
}
\caption{Evaluation of structural metrics for human and model-generated stories under different prompting strategies.}
\label{tab:automatic_metrics_aligned}
\end{table*}

\begin{table*}[t]
  \centering
  \small
  \begin{tabular*}{\textwidth}{@{\extracolsep{\fill}}p{2cm} p{2cm} c c c c c@{}}
    \toprule
    \multicolumn{2}{c}{} & \multicolumn{5}{c}{\textbf{Judge Scores}} \\
    \cmidrule(l){3-7}
    \textbf{Critic} & \textbf{Writer/Editor} & \textbf{Relatedness} & \textbf{\makecell{Creativity}} & \textbf{Fluency} & \textbf{\makecell{Suitability for\\ Children}} & \textbf{Overall} \\ 
    \midrule
    % Baselines
    --- & Human  &$4.69 \pm 0.84$ & $3.20 \pm 0.72$ & $4.55 \pm 0.66$ & $3.95 \pm 1.17$ & $3.83 \pm 0.86$ \\ 
    --- & GPT-4.1 & $4.40 \pm 1.16$ & $3.12 \pm 0.59$ & $4.69 \pm 0.54$ & $4.93 \pm 0.23$ & $3.98 \pm 0.68$ \\
    \midrule
    % Gemma without critic
    --- & Gemma-3-12B & $3.16 \pm 1.60$ & $2.15 \pm 0.44$ & $3.85 \pm 0.63$ & $4.43 \pm 0.70$ & $2.81 \pm 0.69$ \\
    --- & Gemma-3-27B & $3.54 \pm 1.66$ & $2.37 \pm 0.51$ & $4.33 \pm 0.67$ & $4.57 \pm 0.74$ & $3.16 \pm 0.83$ \\
    \cmidrule{3-7}
    % Gemma with critics
    Gemma-3-12B & Gemma-3-12B  &$3.45 \pm 1.63$ & $2.85 \pm 0.66$ & $3.53 \pm 0.61$ & $4.43 \pm 0.82$ & $3.27 \pm 0.79$ \\
    
    Gemma-3-27B & Gemma-3-12B  & $3.56 \pm 1.55$ & $3.02 \pm 0.71$ & $3.58 \pm 0.61$ & $4.43 \pm 0.90$ & $3.31 \pm 0.81$ \\

    \midrule
    Gemma-3-27B & Gemma-3-27B  & $3.75 \pm 1.53$ & $3.18 \pm 0.63$ & $4.21 \pm 0.66$ & $4.78 \pm 0.61$ & $3.68 \pm 0.79$ \\
    
    \bottomrule
  \end{tabular*}

  \caption{LLM-as-a-judge scores (Gemini 2.5 Pro) for stories generated within our feedback-guided semantic decompression experiment.}
  \label{tab:iterative-results}
\end{table*}

\begin{table*}[t]
  \centering
  \small

  \setlength{\tabcolsep}{3pt}
  
  \begin{tabularx}{\textwidth}{l l X X X X X}
    \toprule
    \multicolumn{2}{c}{} & \multicolumn{5}{c}{\textbf{Structural Metrics}} \\
    \cmidrule(l){3-7}
    \textbf{Critic} & \textbf{Writer/Editor} & \textbf{Semantic Diversity} & \textbf{Lexical Diversity} & \textbf{Surprise} & \textbf{Novelty} & \textbf{Readability} \\ 
    \midrule
    % Baselines
    --- & Human
& -- & $0.89 \pm 0.03$ & $0.17 \pm 0.10$ & -- & $92.66 \pm 1.27$ \\

    --- & GPT-4.1           
& $0.38 \pm 0.08$ & $0.91 \pm 0.02$ & $0.16 \pm 0.10$ & $8.17E-2 \pm 0.01$ & $93.30 \pm 0.80$ \\
    \midrule
    % Gemma without critic
    --- & Gemma-3-12B     
& $0.41 \pm 0.07$ & $0.91 \pm 0.02$ & $0.18 \pm 0.10$ & $8.49E-2 \pm 0.01$ & $93.79 \pm 0.85$ \\
    --- & Gemma-3-27B     
& $0.41 \pm 0.07$ & $0.90 \pm 0.02$ & $0.18 \pm 0.09$ & $8.05E-2 \pm 0.01$ & $93.97 \pm 0.61$ \\
    \cmidrule{3-7}
    % Gemma with critics
    Gemma-3-12B & Gemma-3-12B  &$0.43 \pm 0.07$ & $0.89 \pm 0.02$ & $0.19 \pm 0.08$ & $7.35E-2 \pm 0.01$ & $93.45 \pm 0.80$ \\
    
    Gemma-3-27B & Gemma-3-12B  & $0.43 \pm 0.07$ & $0.90 \pm 0.02$ & $0.19 \pm 0.08$ & $ 7.29E-2\pm 0.01$ & $93.75 \pm 0.77$ \\

    \midrule
    Gemma-3-27B & Gemma-3-27B  & $0.42 \pm 0.07$ & $0.90 \pm 0.02$ & $0.21 \pm 0.09$ & $7.19E-2 \pm 0.01$ & $93.77 \pm 0.70$ \\
    
    \bottomrule
   \end{tabularx}

  \caption{Automatic evaluation results for stories generated within our feedback-guided semantic decompression experiment.}
  \label{tab:iterative-results-automatic}
\end{table*}

\section{Guidelines for Annotating Persian Children\'s Stories}
\label{sec:appendixB}
This appendix provides instructions for three graduate annotators who evaluate Persian children’s stories in relation to specific Persian proverbs. Our annotators were Computer Engineering students and did the annotations voluntarily. The annotation process is conducted in Label Studio~\cite{LabelStudio}\footnote{\url{https://labelstud.io/}}, where each story is assessed against five metrics: Relatedness, Creativity, Fluency, Suitability for Children, and Overall. Some stories may have been written by humans, and others may have been generated by large language models; however, all evaluations should be based on the content and the criteria described below.

\subsection*{Evaluation Criteria}
\subsection*{1. Relatedness}
This criterion measures how well a story reflects the meaning of the corresponding proverb.

\begin{itemize}
    \item 1: Not related: No meaningful connection to the proverb.
    \item 2: Weakly related: Only minimal or indirect reference.
    \item 3: Moderately related: Partially conveys the intended meaning.
    \item 4: Well related: Clearly communicates the meaning.
    \item 5: Fully related: Completely and accurately represents the meaning of the proverb.
\end{itemize}
\subsection*{2. Creativity}
This metric assesses how much the story is like a short, creative, or interesting story.

\begin{itemize}
    \item 1: Very uninteresting: Very boring/uninteresting.
    \item 2: Slightly interesting: Limited creativity or entertainment value.
    \item 3: Moderately interesting: Somewhat engaging and creative.
    \item 4: Very interesting: Creative and engaging.
    \item 5: Highly creative: Very creative/story-like
\end{itemize}

\subsection*{3. Fluency}
Fluency assesses the grammatical correctness, clarity, and overall readability of the story.

\begin{itemize}
    \item 1: Very poor: multiple grammatical errors; difficult to read.
    \item 2: Poor: Contains noticeable errors.
    \item 3: Average: Generally readable but includes some issues.
    \item 4: Good: Mostly accurate and fluent.
    \item 5: Excellent: Clear, well-written, and free of errors.
\end{itemize}

\subsection*{4. Suitability for Children}
This criterion determines whether the story’s language, tone, and content are appropriate for a child audience.

\begin{itemize}
    \item 1: Completely unsuitable: Includes themes or elements inappropriate for children.
    \item 2: Slightly unsuitable: Contains issues that reduce child-appropriateness or appeal.
    \item 3: Moderately suitable: Generally acceptable but not strongly oriented toward children.
    \item 4: Suitable: Appropriate, accessible, and appealing for children.
    \item 5: Highly suitable: Strongly child-friendly in tone, vocabulary, and content.
\end{itemize}

\subsection*{5. Overall}
This rating reflects a general evaluation of the overall quality of the story.

\begin{itemize}
    \item 1: Very poor: Lacks coherence or value.
    \item 2: Poor: Contains significant weaknesses.
    \item 3: Average: Acceptable but unremarkable.
    \item 4: Good: Coherent, well-structured, and of solid quality.
    \item 5: Excellent: High-quality.
\end{itemize}

\subsection*{General Annotation Instructions}
\begin{itemize}
    \item \textbf{Objectivity}: Annotators should focus only on the story’s content and how well it aligns with the evaluation criteria. The story’s origin, whether written by a human or generated by a model, should not affect their judgment.
    \item \textbf{Consistency}: Annotators are expected to apply the scoring standards uniformly across all stories to ensure reliability and fairness throughout the dataset.
    \item \textbf{Understanding the Proverb}: Before evaluating any story associated with a proverb, annotators should review the proverb and its intended meaning.
    \item \textbf{Accurate Recording}: All scores ($1$--$5$) must be entered correctly into Label Studio. Annotators should verify that their inputs are properly saved.
    \item \textbf{Optional Notes}: If an annotator wants to comment on a specific annotation or highlight something unusual, they may include it in the comment section provided on the platform.
\end{itemize}

\section{Structural Metric Definitions}
\label{sec:appendix-metrics}

This section provides the formal definitions, equations, and implementation details for all automatic evaluation metrics used in the main paper. It is notable to mention that we have used DadmaTools~\cite{jafari2025dadmatools} for extracting the lemmatization of the tokens for implementing the metrics needed.

\subsection{Diversity}

\paragraph{Lexical Diversity}

Lexical diversity measures surface-level variation in word usage. It is defined as the ratio of unique $n$-grams to the total number of $n$-grams in a story, for $n \in \{1,2,3,4,5\}$:
\[
\text{LexDiv}_n =
\frac{\#\text{unique } n\text{-grams}}{\#\text{total } n\text{-grams}}.
\]

Values closer to 1 indicate richer and less repetitive language.

\paragraph{Semantic Diversity}

Semantic diversity captures variation in meaning across stories written for the same proverb. Following~\cite{padmakumar2023does}, we use the \emph{inverse homogenization score}, defined as the average pairwise semantic distance between stories for a given proverb.

Given a story $s$ written for proverb $p$, the score is defined as:
\[
\text{inv\_hom}(s \mid p) =
\frac{1}{|S_p| - 1}
\sum_{s' \neq s}
\text{semdis}(s, s'),
\]
where $S_p$ is the set of all stories written for proverb $p$. In our experiments, $|S_p| = 2$, consisting of one human-written story and one LLM-generated story. The semantic distance between two stories, ${semdis}(s, s’)$, is defined as $1 -{cosine\_similarity}$, where cosine similarity is computed between sentence embeddings. Sentence embeddings are generated using the \textit{bge-m3} model~\cite{chen2024bge}. The resulting semantic distance lies in the range $[0, 2]$, where higher values indicate greater semantic divergence.

\subsection{Novelty}

Novelty measures how semantically different a story is relative to the overall corpus~\cite{RuncoJaeger2012, Maher2010}. Following~\cite{Karampiperis2014,johnson2023divergent}, each story is represented using dominant terms, defined as lemmatized content words.

Let $S_n$ denote a story and $S_G$ denote the corpus of all stories, with dominant term sets $T_n$ and $T_G$, respectively. We compute the average semantic distance between all unordered term pairs within a story:
\[
D(S_n) =
\frac{1}{|P(T_n)|}
\sum_{(i,j) \in P(T_n)}
\text{semdis}(T_{n,i}, T_{n,j}),
\]
and analogously for the corpus:
\[
D(S_G) =
\frac{1}{|P(T_G)|}
\sum_{(i,j) \in P(T_G)}
\text{semdis}(T_{G,i}, T_{G,j}).
\]

The novelty of the story $S_n$ is defined as:
\[
\text{Nov}(S_n) =
2 \cdot \left| D(S_n) - D(S_G) \right|.
\]

The novelty score ranges from 0 to 2, where higher values indicate greater semantic deviation from the corpus.

\subsection{Surprise}

Surprise reflects how much a story deviates from expectations as it unfolds~\cite{Boden1991Creative, Maher2010, GraceMaher2014}. Following~\cite{Karampiperis2014}, surprise is computed by comparing consecutive sentences.

Let a story $S_n$ consist of $|F|$ sentences $\{F_1, \dots, F_{|F|}\}$. The surprise score is defined as:
\[
\text{Sur}(S_n) =
\frac{2}{|F| - 1}
\sum_{i=2}^{|F|}
\left| D(F_i) - D(F_{i-1}) \right|,
\]
where $D(\cdot)$ refers to the average semantic distance defined in the previous metric. This score lies in the range $[0,2]$. Lower values indicate gradual semantic changes between sentences, while higher values indicate stronger and more unexpected shifts in meaning.

\subsection{Readability}

To assess readability for Persian texts, we employ the Flesch--Dayani metric~\cite{dayani2000criteria}, which estimates reading difficulty based on surface linguistic features:
\[
\text{FD} =
262.835
- 0.846 \cdot \frac{|\text{letters}|}{|\text{words}|}
- 1.01 \cdot \frac{|\text{words}|}{|\text{sentences}|}.
\]

Higher Flesch--Dayani scores indicate easier-to-read text, while lower scores correspond to greater textual complexity.

\section{Statistical Alignment of Feedback-Guided Enhancements with Human Preference}
\label{sec:metric_correlations}
To establish a human ground truth for quality improvement, we employed the same three expert annotators who collaborated on collecting the gold data for our subjective metrics. We presented them with 40 randomly selected pairs of initial and final story drafts from the Gemma 3 27B self-correction experiment in a completely blind process (i.e., without revealing which version was the refined draft). Annotators were asked to select the preferred story.

As shown in Table~\ref{tab:human_preference}, results of the human evaluation indicate a consistent preference for the post-feedback final drafts over the initial drafts across all three independent annotators. Individually, Annotators 1, 2, and 3 favored the final stories in 29, 23, and 25 of the evaluated pairs, respectively. When these individual selections are aggregated via majority vote, the final drafts were chosen 65\% of the time (26 out of 40 pairs), compared to only 35\% (14 out of 40) for the initial drafts. Furthermore, out of the 40 samples, all three annotators were perfectly unanimous on 15 stories. Notably, 13 of these 15 exact matches favored the post-feedback stories, with only 2 favoring the initial drafts. This clear majority and the number of full consensus demonstrate that the applied feedback loop consistently yields noticeable improvements in the overall quality and appeal of the generated stories according to human judges.  

\begin{table}[th]
\centering
\small
\renewcommand{\arraystretch}{1.2}
\resizebox{\columnwidth}{!}{%
\begin{tabular}{l c c}
    \toprule
    \textbf{Annotator} & 
    \textbf{\makecell{Initial Draft\\(before feedback)\\Selected}} & 
    \textbf{\makecell{Final Draft\\(after feedback)\\Selected}} \\
    \midrule
    Annotator 1 & 11 & 29 \\
    Annotator 2 & 17 & 23 \\
    Annotator 3 & 15 & 25 \\
    \midrule
    \makecell[l]{Majority Vote\\between annotators} & 14 (35\%) & 26 (65\%) \\
    \bottomrule
\end{tabular}%
}
\caption{Results of a blinded human pairwise preference study on 40 story pairs.}
\label{tab:human_preference}
\end{table}

To evaluate the alignment between human preferences and our calibrated LLM-as-a-Judge (Gemini 2.5 pro) scores, we analyzed the correlation between human choices and the change in our subjective metric scores following the refinement loop in our feedback-guided decompression regime. Given the limited sample size ($N=40$) of the human evaluation set, we employed a non-parametric bootstrap approach to estimate the robustness of the observed correlations.

\begin{table}[th]
    \centering

    \resizebox{\columnwidth}{!}{%
        \begin{tabular}{lccc}
            \toprule
            \textbf{Metric} & \textbf{Pearson $r$} & \textbf{95\% CI} & \textbf{Sig.} \\
            \midrule
            Creativity & 0.24 & $[-0.09, 0.54]$ & No \\
            Relatedness & 0.19 & $[-0.06, 0.42]$ & No \\
            Overall Score & 0.17 & $[-0.15, 0.45]$ & No \\
            Suitability (Children) & 0.08 & $[-0.25, 0.40]$ & No \\
            Fluency & -0.04 & $[-0.37, 0.30]$ & No \\
            \bottomrule
        \end{tabular}%
    }
        \caption{Bootstrap analysis ($N=40$, 10k iterations) of the correlation between metric score improvements and human preference. The table is sorted by correlation strength.}
    \label{tab:metric_correlations}
\end{table}

We performed 10,000 bootstrap iterations. In each iteration, we resampled the 40 data pairs (human preference, metric change) with replacement and calculated the Pearson correlation coefficient ($r$). The reported Mean Correlation represents the average $r$ across all iterations. To assess statistical significance, we computed 95\% Confidence Intervals (CI) using the percentile method (2.5th and 97.5th percentiles of the bootstrap distribution).

Table~\ref{tab:metric_correlations} summarizes the results. A metric is considered to have a statistically significant correlation with human preference at the $\alpha=0.05$ level if the 95\% CI does not cross zero. As shown in the table, while \textit{Creativity} and \textit{Relatedness} exhibit positive directional trends (mean $r > 0.19$), the confidence intervals for all metrics include zero, indicating that the correlations are not statistically significant for this sample size.

\section{Examples of Decompression Gap}
\label{sec:appendix-examples}
In this appendix section, we highlight instances of the decompression gap, illustrating situations where language models generate highly fluent text but fail to accurately ground abstract concepts into their narratives. Table~\ref{tab:generated_stories_eval} presents examples of this phenomenon using GPT 4.1 instructed by the Moral CoT prompt to write stories based on proverbs. In these cases, the model ignores the underlying moral intent of the idioms and instead builds stories around their literal interpretations. We attribute this shortcoming to the inherent difficulty of constrained semantic decompression, alongside a noticeable deficit in cultural knowledge and alignment within the model.

For example, the literal translation of the first proverb in Table~\ref{tab:generated_stories_eval} is "What you read from the page, I know by heart." Its true cultural meaning emphasizes that an experienced person possesses deep, internalized wisdom and does not need to rely on explicit sources. Rather than capturing this contrast between superficial knowledge and true mastery, GPT 4.1 took the phrase literally and wrote a story specifically about a child memorizing a school poem, completely missing the deeper theme of wisdom.

The second proverb translates literally to "Neither a khan has come, nor a khan has gone." This idiom is culturally used to express that a situation remains entirely unchanged and that someone's actions had absolutely no impact. Once more, GPT 4.1 failed to build a narrative embodying this concept. Instead, it simply dropped the literal phrase into the dialogue of an unrelated story about a stolen seed.

Finally, the third proverb translates to "With a new sleeve, eat the rice." Culturally, this is a cynical observation about how society values outward appearance and wealth over true inner worth. However, GPT 4.1 generated a story that concluded with the exact opposite moral. It wrote a narrative teaching that inner character matters more than clothes, thereby entirely reversing the intended cynical message of the original proverb.

\begin{table*}[th!]
\centering
\scriptsize
\renewcommand{\arraystretch}{1.1}
\setlength{\tabcolsep}{3pt} 

\definecolor{translationgray}{gray}{0.45}
\newcommand{\transl}[1]{{\color{translationgray}\textit{#1}}}

\begin{tabularx}{\textwidth}{c p{3cm} X c c}
    \toprule
    \textbf{Idx} & \textbf{Proverb} & \textbf{Generated Story} & \textbf{Rel.} & \textbf{Flu.} \\
    \midrule
    
    1 & 
    \begin{farsi}
    آنچه تو از رو می خوانی، من از بَرَم
    \end{farsi}

    \transl{(What you read from the page, I know by heart.)} & 
    \begin{farsi}
    در دهکده‌ای سبز و آرام، دو دوست به نام‌های سحر و نسترن زندگی می‌کردند. سحر خیلی باهوش بود و عاشق یادگیری شعرهای جدید، اما نسترن همیشه دوست داشت حرف‌های تازه را از روی کتاب بخواند.
    
    \vspace{0.2em}
    
    یک روز مربی مدرسه از بچه‌ها خواست شعری را حفظ کنند و فردا برایش بخوانند. شب، سحر با شور و شوق شعر را چند بار خواند و آن را به راحتی از بر کرد اما نسترن نتوانست شعر را حفظ کند و هر بار مجبور شد به کتاب نگاه کند.
    
    \vspace{0.2em}
    
    فردا صبح، بچه‌ها جمع شدند. نخست نوبت نسترن شد. او کتاب به دست گرفت و سعی کرد شعر را از روی کتاب بخواند اما بعضی بیت‌ها را اشتباه خواند. بعد نوبت سحر رسید. او با لبخند و صدایی بلند، شعر را بدون هیچ نگاه کردنی به کتاب، کامل و درست خواند.
    
    \vspace{0.2em}
    
    بچه‌ها تشویقش کردند. مربی مهربان گفت: "بعضی‌ها مثل سحر هستند که زود چیزها را یاد می‌گیرند و از بر می‌کنند، بعضی‌ها هم باید بیشتر تمرین کنند تا به همان خوبی شوند. هرکسی راه خودش را دارد، اما همیشه تلاش و تمرین از همه مهم‌تر است."
    
    \vspace{0.2em}
    
    از آن روز به بعد، نسترن از سحر خواست که راه‌های خوب حفظ کردن شعر را به او یاد بدهد. سحر با خوشحالی به دوستی‌اش ادامه داد و با هم شعرهای بیشتری یاد گرفتند و دیگر تنها به خواندن از روی کتاب قانع نشدند.
    \end{farsi}

    \vspace{0.3em}
    \transl{(In a calm and green village, two friends named Sahar and Nastaran lived together. Sahar was very clever and loved learning new poems, but Nastaran always preferred to read new things from a book. One day, their teacher asked the children to memorize a poem and recite it the next day. That night, Sahar eagerly read the poem several times and easily memorized it, while Nastaran could not and had to keep looking at the book. The next morning, the children gathered. Nastaran went first. She held the book and tried to read the poem from it, but she misread some lines. Then it was Sahar's turn. She smiled and recited the poem aloud, perfectly, without glancing at the book. The children cheered. The kind teacher said: ``Some people, like Sahar, learn things quickly and memorize them; others need more practice to reach the same level. Everyone has their own path, but effort and practice are always what matter most.'' From that day on, Nastaran asked Sahar to teach her good ways to memorize poems. Sahar happily continued their friendship, and together they learned more poems, no longer content with only reading from the book.)} &
    2 & 5 \\
    \midrule
    
    2 & 
    \begin{farsi}
    نه خانی آمده، نه خانی رفته
    \end{farsi}

    \transl{(Neither has a khan come, nor has a khan gone.)} & 
    \begin{farsi}
    در یک دهکده‌ی کوچک و زیبا، تعدادی مرغ و خروس در کنار هم زندگی می‌کردند. آن‌ها هر روز صبح با هم بیدار می‌شدند و دنبال دانه و کرم می‌گشتند. میان گله، یک خروس پُر سر و صدا به اسم قلقلی بود که همیشه دوست داشت ماجرا درست کند.

    \vspace{0.2em}
    
    یک روز، وقتی همه مرغ‌ها و خروس‌ها مشغول خوردن بودند، قلقلی چشمش به یک دانه‌ی بزرگ افتاد که کنار لانه افتاده بود. دانه را برداشت و با خودش فکر کرد: «اگر این دانه را خودم تنهایی بخورم، هیچ کس هم متوجه نمی‌شود!» او دانه را قورت داد و آرام به کنار دیگر مرغ‌ها برگشت.

    \vspace{0.2em}
    
    ولی همان موقع، خروس دانا که اسمش مش قُمری بود با صدای بلند پرسید: «بچه‌ها! کسی دانه‌ی بزرگ کنار لانه را دیده؟ من دلم می‌خواست برای همه از آن دانه نقل درست کنم.» مرغ‌ها و خروس‌ها با تعجب به هم نگاه کردند. هیچ‌کس خبر نداشت دانه کجا رفته... ولی قلقلی با معصومیت شانه بالا انداخت و طوری رفتار کرد که انگار از چیزی خبر ندارد.

    \vspace{0.2em}
    
    مش قُمری کمی این طرف و آن طرف رفت، با کله‌اش خاک‌ها را کنار زد اما اثری از دانه پیدا نکرد. بالاخره گفت: «عجب! پس دانه گم شد!» مرغ‌ها کمی پچ‌پچ کردند ولی بعد همه فراموش کردند و دوباره مشغول غذا خوردن شدند.

    \vspace{0.2em}
    
    روز گذشت و هیچ‌کس راجع به دانه حرفی نزد. قلقلی با خودش فکر کرد: «عجب! نه خانی آمده، نه خانی رفته! کسی نفهمید چه اتفاقی افتاد!» اما شب وقتی در لانه خوابیده بود، کمی دلش گرفت. با خودش فکر کرد: «اگر دانه را به دوستانم می‌دادم، همه شاد می‌شدیم و با هم می‌خندیدیم، ولی اکنون فقط ساکت و بی‌صدا یک دانه خوردم.»

    \vspace{0.2em}
    
    صبح فردا، قلقلی دلش طاقت نیاورد و پیش مش قُمری رفت. با شرمندگی گفت: «داداش قمری، من دانه را خوردم و چیزی نگفتم. دلم می‌خواست همه خوشحال شوند، اما خودخواهی کردم. دیگر تکرار نمی‌کنم.»

    \vspace{0.2em}
    
    مش قُمری لبخند زد و گفت: «مهم این است که راستش را گفتی و دوست داشتی شادی را تقسیم کنی. ببین چطور همه چیز دوباره خوب شد.»

    \vspace{0.2em}
    
    از آن روز به بعد، قلقلی یاد گرفت اگر اشتباهی کرد و نادیده گرفت، شاید دیگران متوجه نشوند، اما خودش بهتر از هر کسی می‌فهمد و بهتر است شجاع باشد و راستش را بگوید.
    \end{farsi}

    \vspace{0.3em}
    \transl{(In a small and beautiful village, a group of hens and roosters lived together. Every morning, they woke up together and searched for seeds and worms. Among the flock was a very noisy rooster named Gholgholi, who always loved to stir up trouble. One day, while all the hens and roosters were busy eating, Gholgholi noticed a large seed lying beside the coop. He picked it up and thought to himself, “If I eat this seed all by myself, no one will even notice!” He swallowed the seed and quietly returned to the others. Just then, the wise rooster, whose name was Mash Ghomri, loudly asked, “Everyone! Has anyone seen the big seed beside the coop? I wanted to make a sweet treat from it for all of us.” The hens and roosters looked at each other in surprise. No one knew where the seed had gone… but Gholgholi innocently shrugged his shoulders and acted as if he knew nothing about it. Mash Ghomri searched here and there, pushing the dirt aside with his beak, but he could not find any sign of the seed. Finally, he said, “How strange! The seed has disappeared!” The hens whispered among themselves for a while, but soon everyone forgot about it and went back to eating. The day passed, and no one talked about the seed again. Gholgholi thought to himself, “Well, neither has a khan come, nor has a khan gone! No one realized what I did!” But that night, as he lay in the coop, he began to feel sad. He thought, “If I had shared the seed with my friends, we all could have been happy and laughed together. But instead, I selfishly ate it all alone.” The next morning, Gholgholi could no longer bear the guilt and went to Mash Ghomri. Embarrassed, he said, “Brother Ghomri, I ate the seed and didn’t say anything. I wanted everyone to be happy, but I acted selfishly. I won't do it again.” Mash Ghomri smiled and said, “What matters is that you told the truth and wanted to share happiness with others. See how everything became good again.” From that day on, Gholgholi learned that if he made a mistake and ignored it, others might not notice, but he himself would always know. And it is better to be brave and tell the truth.)} &
    2 & 5 \\
    \midrule
    
    3 & 
    \begin{farsi}
    آستین نو پلو بخور
    \end{farsi}

    \transl{(With a new sleeve, eat rice.)} & 
    \begin{farsi}
    روزی روزگاری، موش کوچولویی به اسم نازی بود که لباس‌هایش کهنه و پاره بود. روزی شنید قرار است در جشن جنگل غذای خوشمزه‌ای بدهند اما فقط به مهمان‌های خوش‌لباس غذا می‌دهند. نازی ناراحت شد، تا اینکه روباه مهربان یک آستین نوی زیبا به او داد. نازی با خوشحالی آستین را پوشید و به جشن رفت. اما او به جز آستین، همان لباس‌های کهنه‌ی همیشگی را داشت.

    \vspace{0.2em}
    
    وقتی نوبت غذا خوردن او رسید، میزبان به آستین او نگاه کرد و با تعجب گفت: "چرا فقط آستینت نو است؟ تو همانی که همیشه مهربان و خوش‌رفتاری، ما تو را به خاطر دل خوبت دوست داریم، نه لباست!" نازی یاد گرفت که ارزش آدم‌ها به دل و رفتارشان است، نه لباس‌هایشان. از آن روز نازی با امید بیشتری زندگی کرد و فهمید دوستی و مهربانی از هر لباسی زیباتر است.
    \end{farsi}

    \vspace{0.3em}
    \transl{(Once upon a time, there was a little mouse named Nazi whose clothes were old and torn. One day she heard that at the forest festival delicious food would be served, but only to well-dressed guests. Nazi felt sad, until a kind fox gave her a beautiful new sleeve. Happily, Nazi put it on and went to the festival. But apart from the sleeve, she was wearing the same old clothes as always. When it was her turn to eat, the host looked at her sleeve and said in surprise: ``Why is only your sleeve new? You are the one who is always kind and well-mannered. We love you for your good heart, not your clothes!'' Nazi learned that a person's worth lies in their heart and behavior, not their clothing. From that day on, Nazi lived with greater hope and understood that friendship and kindness are more beautiful than any clothing.)} &
    1 & 4 \\
    \bottomrule
\end{tabularx}
\vspace{-1.0em}
\caption{Examples of decompression gap for generated stories by GPT 4.1 using the Moral CoT prompt. Literal English translations are provided in parentheses in gray below the original Persian text. These translations are not included in the prompts provided to the model or in the model's outputs. Rel.\ = Judge Relatedness Score, Flu.\ = Judge Fluency Score.}
\label{tab:generated_stories_eval}
\end{table*}

\section{Prompt Templates}
\begin{figure*}[t] 
    \centering
    \includegraphics[width=1.0\textwidth]{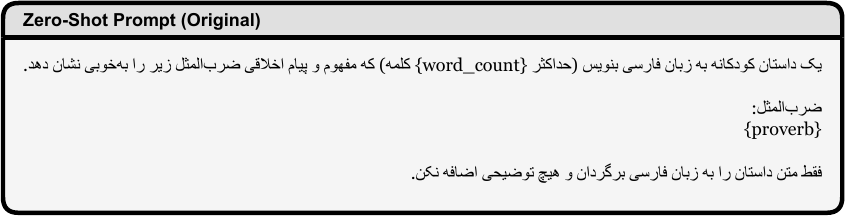} 
    \caption{Zero-shot story generation prompt (original Persian).}
    \label{fig:persian-zeroshot}
\end{figure*}

\begin{figure*}[t]
    \centering
    \includegraphics[width=1.0\textwidth]{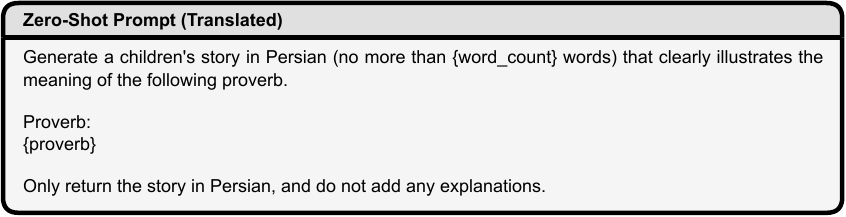} 
    \caption{Zero-shot story generation prompt (translated from Persian).}
    \label{fig:english-zeroshot}
\end{figure*}

\begin{figure*}[t]
    \centering
    \includegraphics[width=1.0\textwidth]{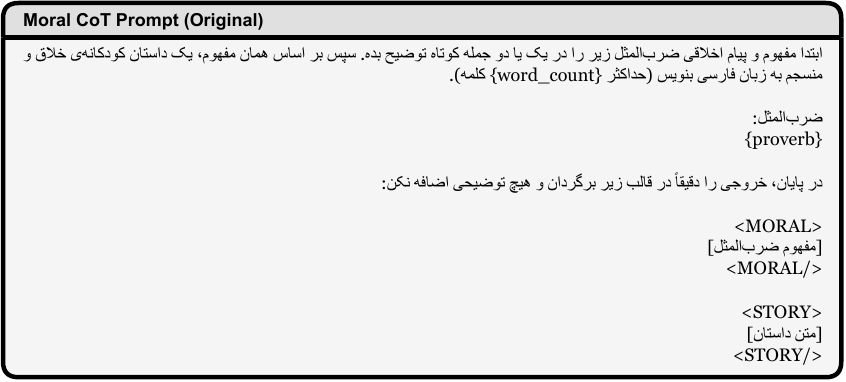} 
    \caption{Moral CoT story generation prompt (original Persian).}
    \label{fig:persian-cot-moral}
\end{figure*}

\begin{figure*}[t] 
    \centering
    \includegraphics[width=1.0\textwidth]{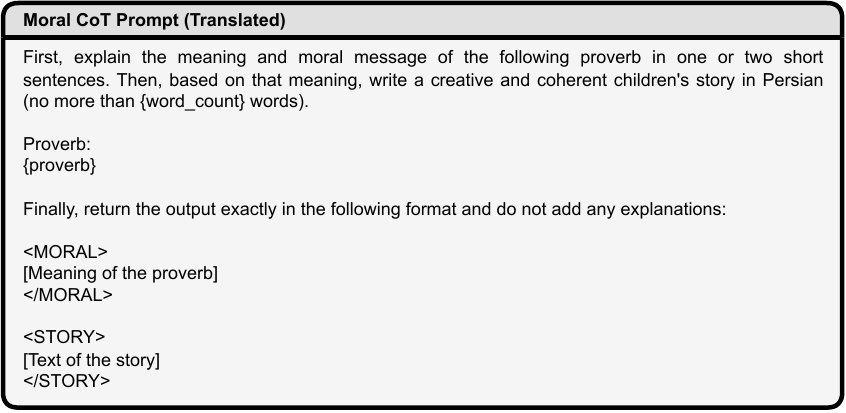} 
    \caption{Moral CoT story generation prompt (translated from Persian).}
    \label{fig:english-cot-moral}
\end{figure*}

\begin{figure*}[t] 
    \centering
    \includegraphics[width=1.0\textwidth]{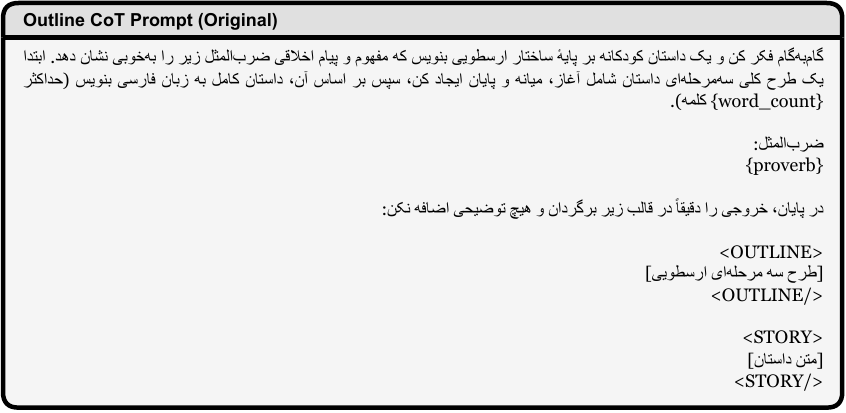} 
    \caption{Outline  CoT story generation prompt (original Persian).}
    \label{fig:persian-cot-outline}
\end{figure*}

\begin{figure*}[t] 
    \centering
    \includegraphics[width=1.0\textwidth]{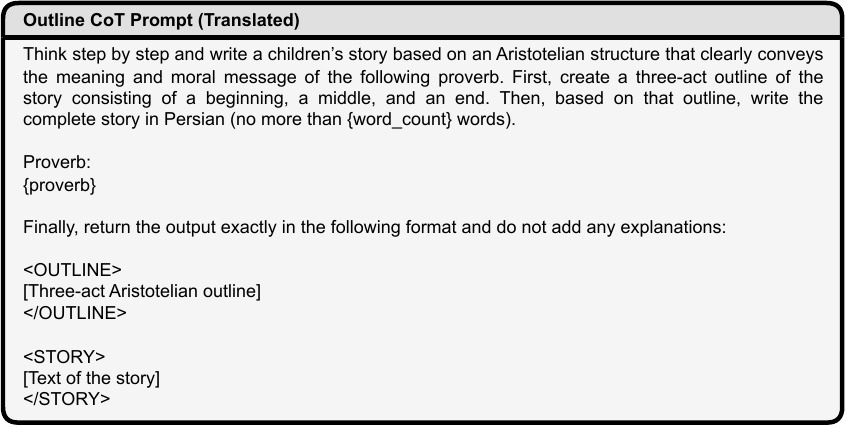} 
    \caption{Outline  CoT story generation prompt (translated from Persian).}
    \label{fig:english-cot-outline}
\end{figure*}

\begin{figure*}[t] 
    \centering
    \includegraphics[width=1.0\textwidth]{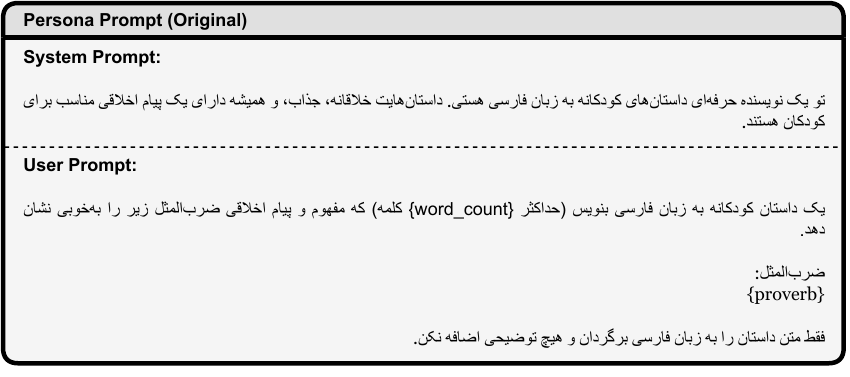} 
    \caption{Persona story generation prompt (original Persian).}
    \label{fig:persian-persona}
\end{figure*}

\begin{figure*}[t]
    \centering
    \includegraphics[width=1.0\textwidth]{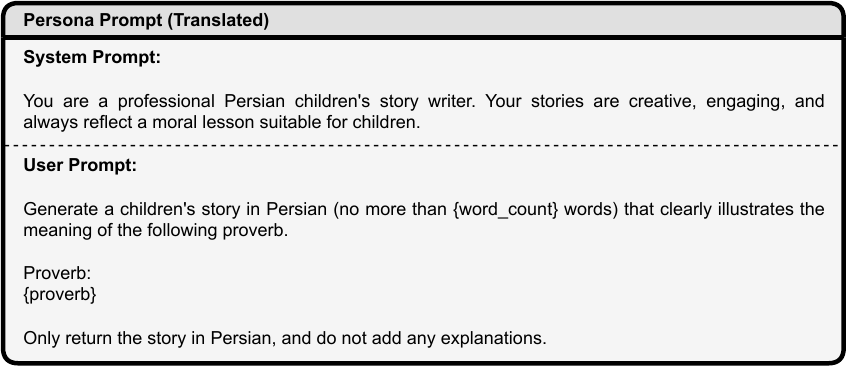} 
    \caption{Persona story generation prompt (translated from Persian).}
    \label{fig:english-persona}
\end{figure*}

\begin{figure*}[t] 
    \centering
    \includegraphics[width=1.0\textwidth]{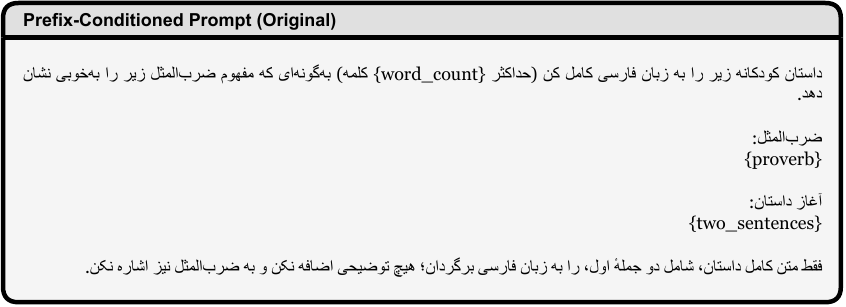} 
    \caption{Prefix-Conditioned story generation prompt (original Persian).}
    \label{fig:persian-prefix}
\end{figure*}

\begin{figure*}[t]
    \centering
    \includegraphics[width=1.0\textwidth]{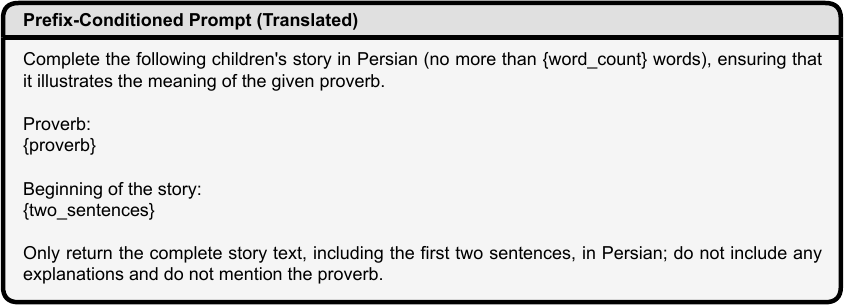} 
    \caption{Prefix-Conditioned story generation prompt (translated from Persian).}
    \label{fig:english-prefix}
\end{figure*}

\begin{figure*}[t] 
    \centering
    \includegraphics[width=1.0\textwidth]{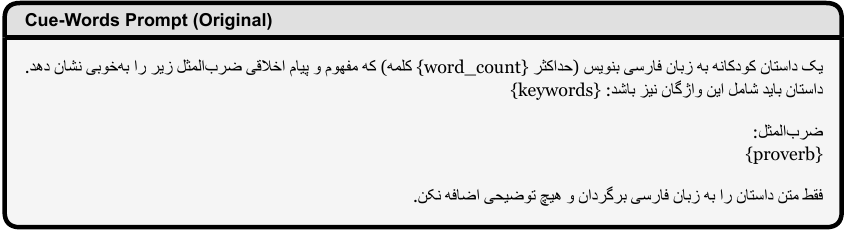} 
    \caption{Cue-Words story generation prompt (original Persian).}
    \label{fig:persian-keywords}
\end{figure*}

\begin{figure*}[t] 
    \centering
    \includegraphics[width=1.0\textwidth]{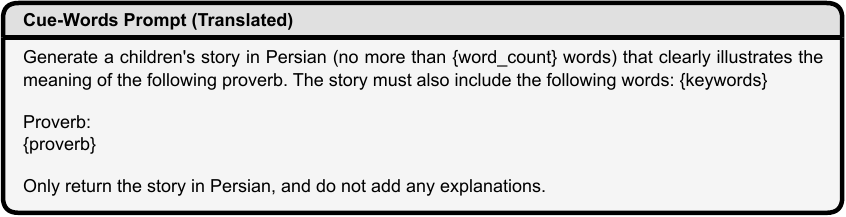} 
    \caption{Cue-Words story generation prompt (translated from Persian).}
    \label{fig:english-keywords}
\end{figure*}

\begin{figure*}[t] 
    \centering
    \includegraphics[width=1.0\textwidth]{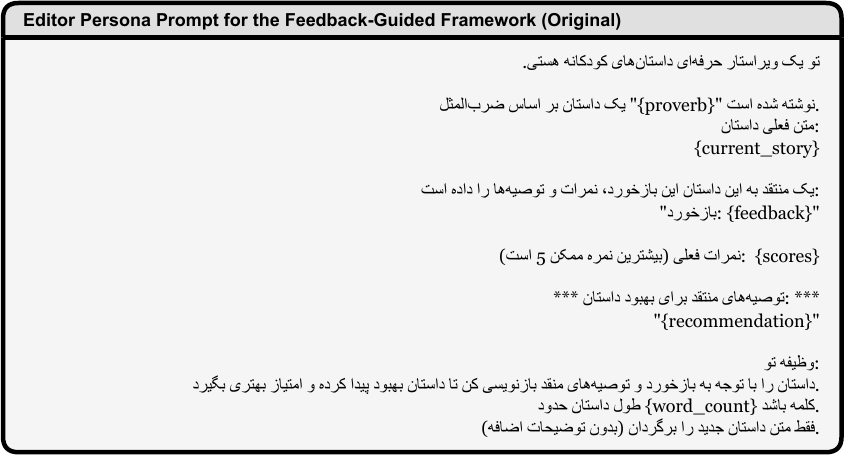} 
    \caption{Editor persona prompt for the feedback-guided framework (original Persian).}
    \label{fig:persian-editor}
\end{figure*}

\begin{figure*}[t] 
    \centering
    \includegraphics[width=1.0\textwidth]{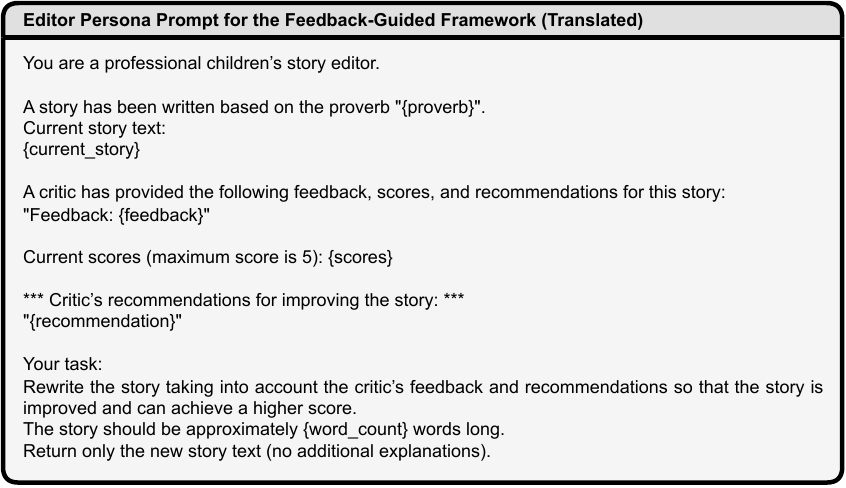} 
    \caption{Editor persona prompt for the feedback-guided framework (translated from Persian).}
    \label{fig:english-editor}
\end{figure*}

\begin{figure*}[t]
    \centering
    \includegraphics[width=1.0\textwidth]{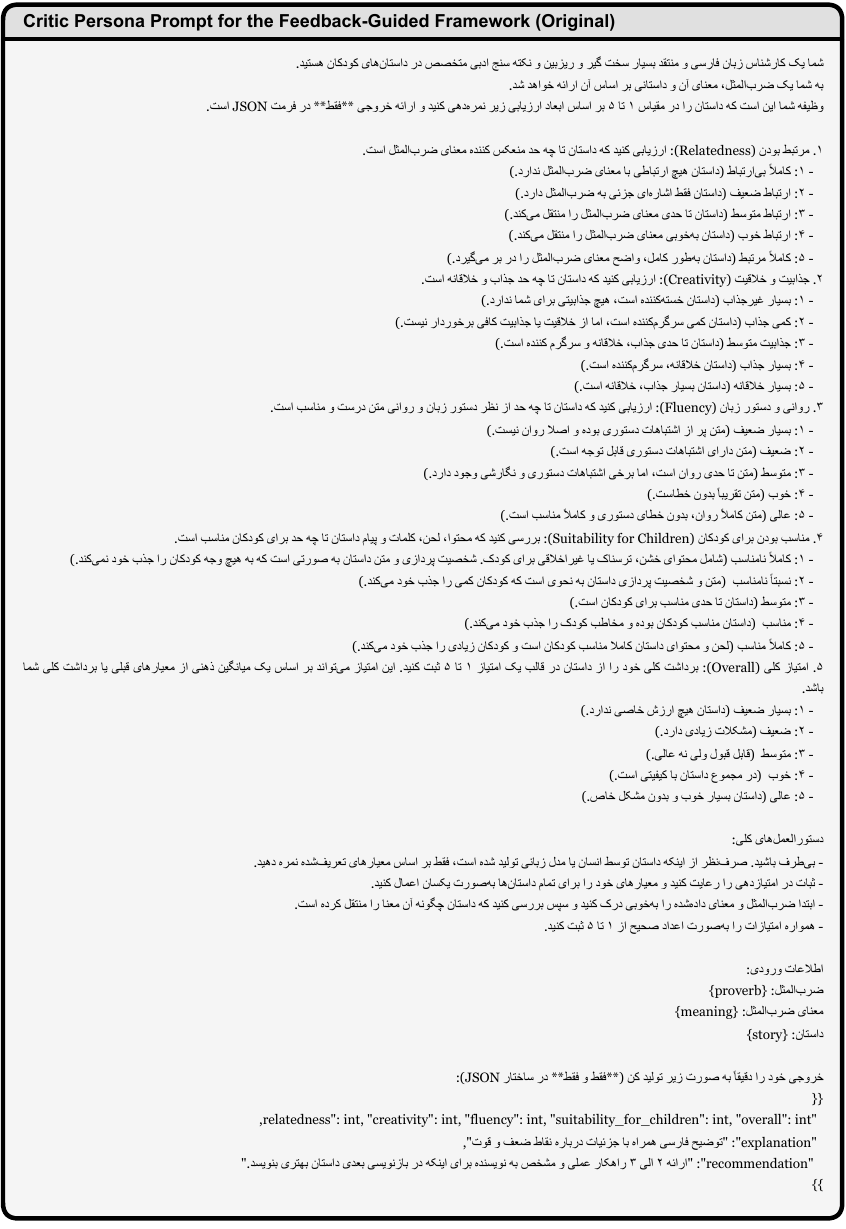} 
    \caption{Critic persona prompt (original Persian) for the feedback-guided framework. As Gemma models lack system role support \cite{gemmateam2025gemma3technicalreport}, the persona is defined within the user prompt.}
    \label{fig:persian-critic}
\end{figure*}

\begin{figure*}[t]
    \centering
    \includegraphics[width=1.0\textwidth]{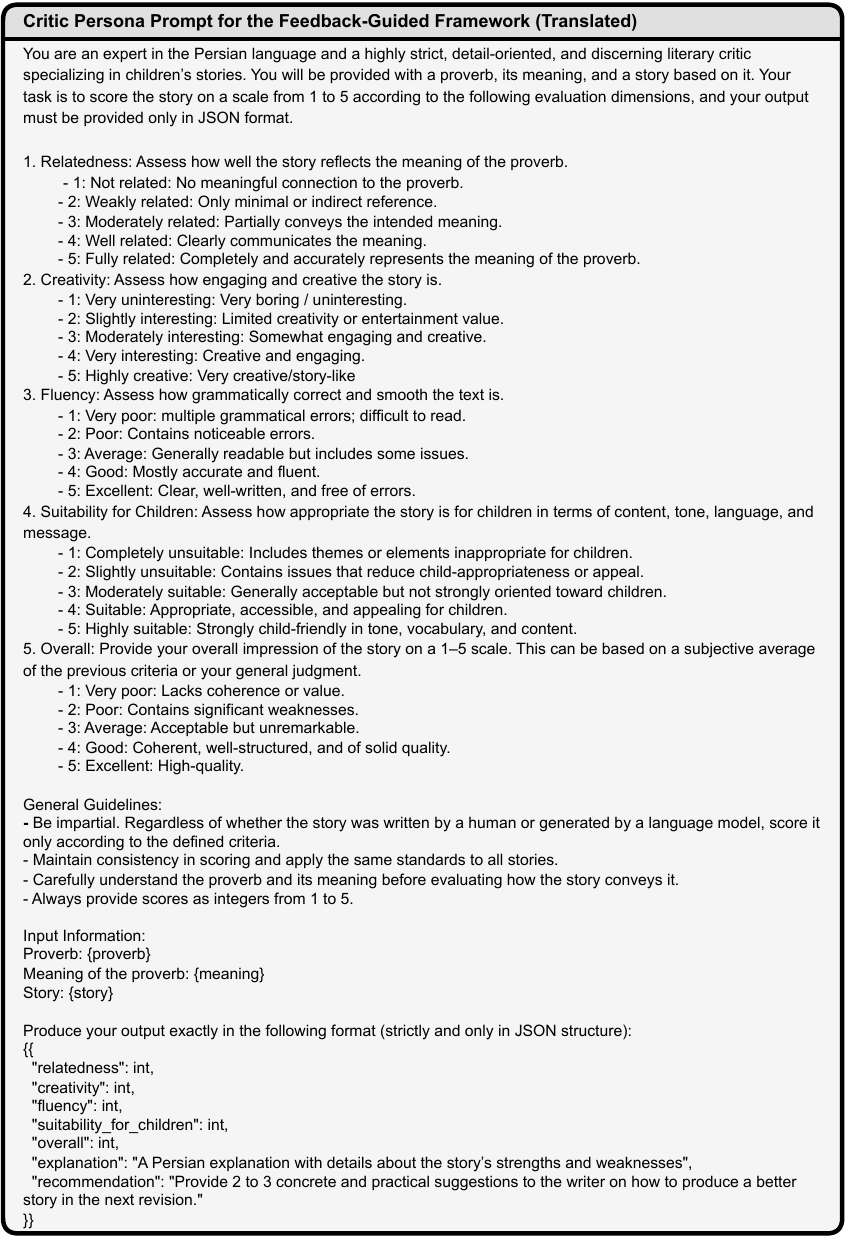}
    \vspace{-2em}
    \caption{Critic persona prompt (translated from Persian) for the feedback-guided framework. As Gemma models lack system role support \cite{gemmateam2025gemma3technicalreport}, the persona is defined within the user prompt.}
    \label{fig:english-critic}
\end{figure*}

\begin{figure*}[t] 
    \centering
    \includegraphics[width=1.0\textwidth]{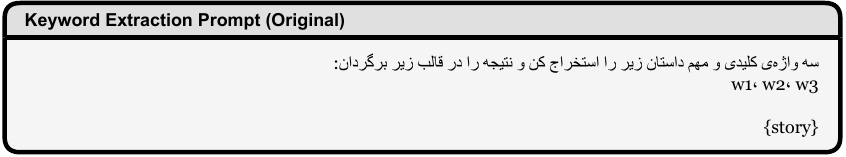} 
    \caption{Keyword extraction prompt (original Persian).}
    \label{fig:persian-cue}
\end{figure*}

\begin{figure*}[t] 
    \centering
    \includegraphics[width=1.0\textwidth]{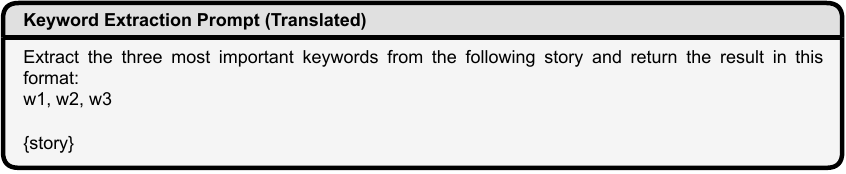} 
    \caption{Keyword extraction prompt (translated from Persian).}
    \label{fig:english-cue}
\end{figure*}

\begin{figure*}[t]
    \centering
    \includegraphics[width=1.0\textwidth]{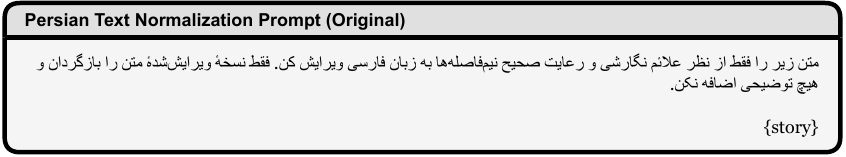} 
    \caption{Punctuation and half-space normalization prompt (original Persian).}
    \label{fig:persian-edit}
\end{figure*}

\begin{figure*}[t]
    \centering
    \includegraphics[width=1.0\textwidth]{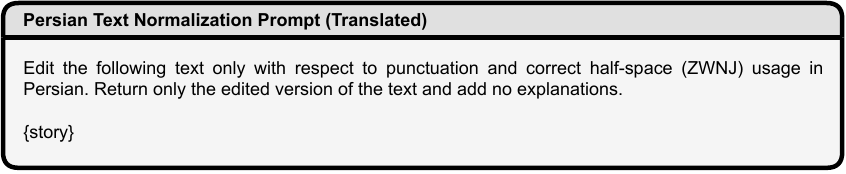} 
    \caption{Punctuation and half-space normalization prompt (translated from Persian).}
    \label{fig:english-edit}
\end{figure*}

\begin{figure*}[t] 
    \centering
    \includegraphics[width=1.0\textwidth]{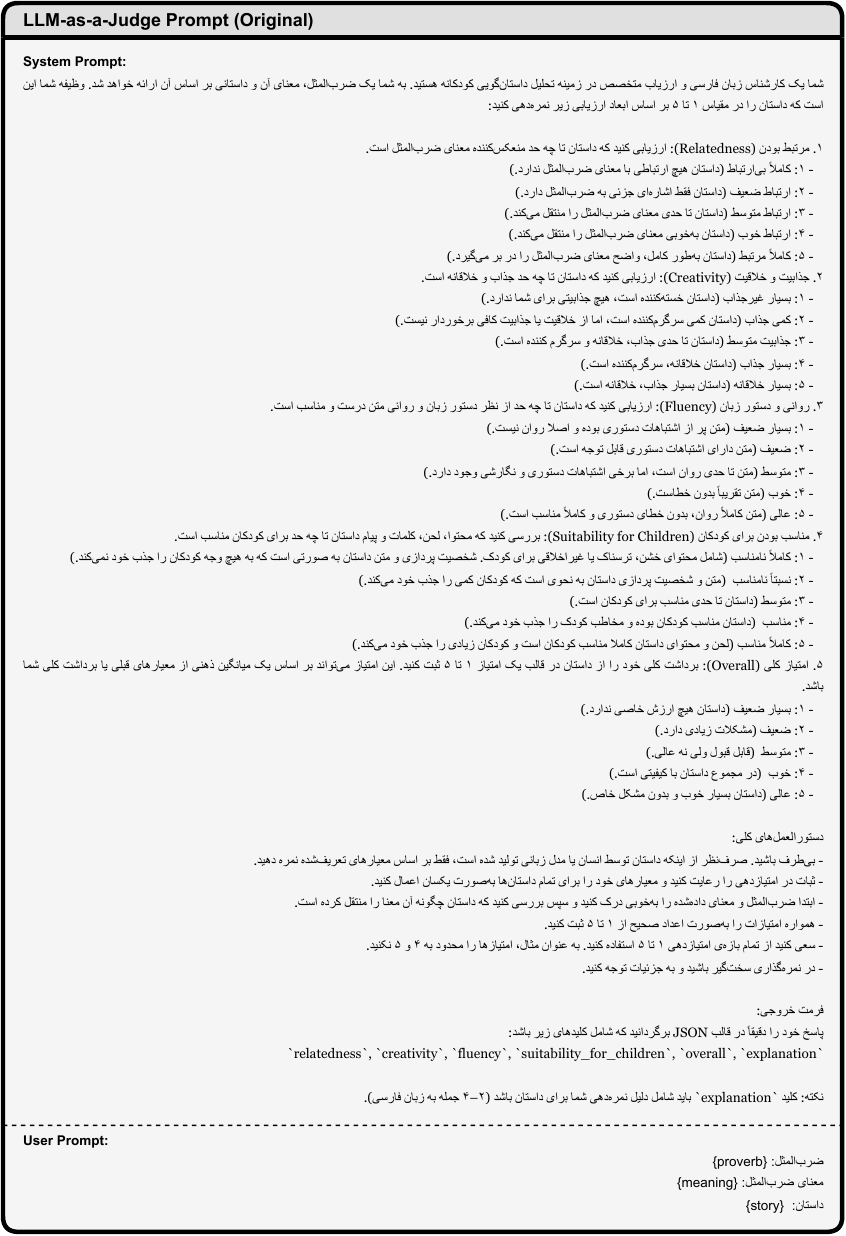} 
    \caption{LLM-as-a-judge prompt (original Persian).}
    \label{fig:persian-judge}
\end{figure*}

\begin{figure*}[t] 
    \centering
    \includegraphics[width=1.0\textwidth]{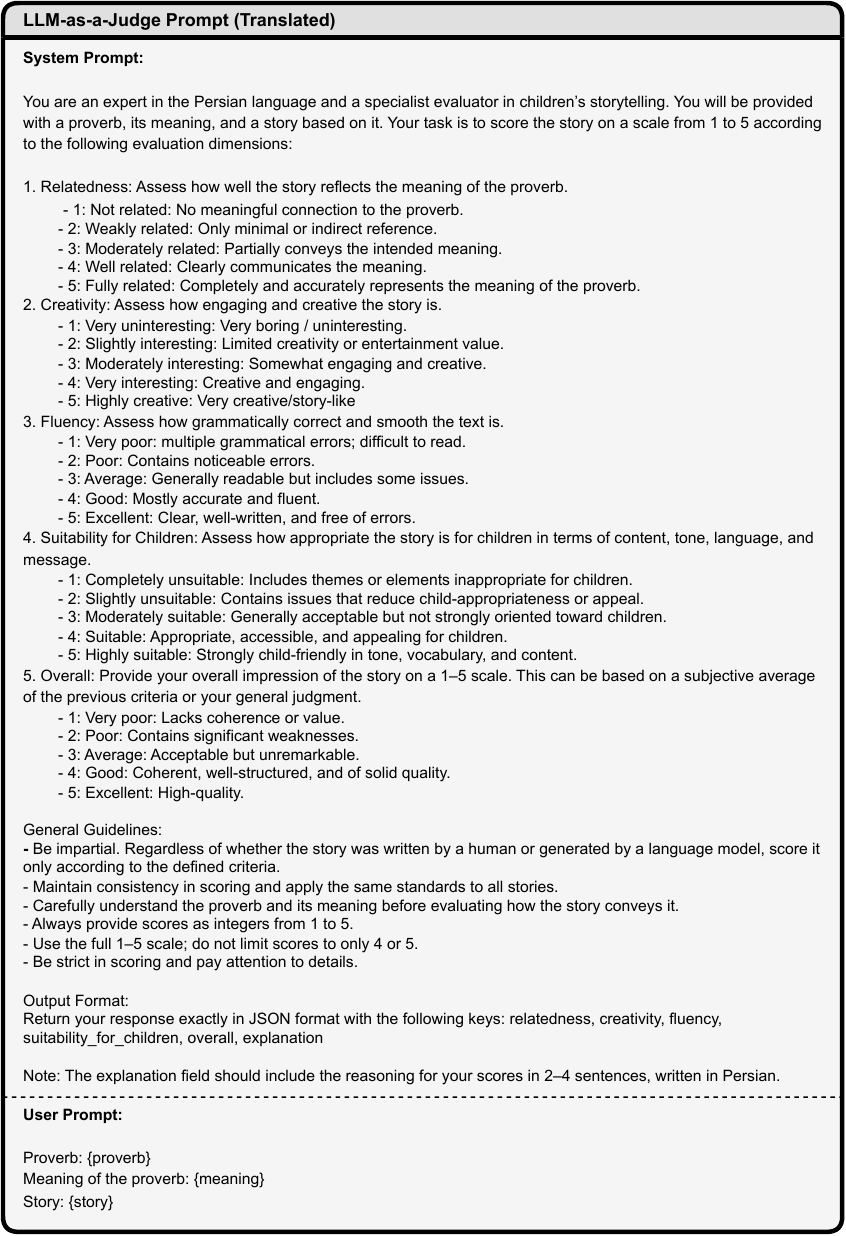} 
    \caption{LLM-as-a-judge prompt (translated from Persian).}
    \label{fig:english-judge}
\end{figure*}

All prompt templates used in our experiments are presented in this appendix, in both their original Persian form and their English translations, spanning Figures \ref{fig:persian-zeroshot} through \ref{fig:english-judge}.
\label{sec:appendix-prompt}

\FloatBarrier

\end{document}